\documentclass[10pt,journal,compsoc]{IEEEtran}
\usepackage{amsfonts}

\usepackage{array}
\usepackage[caption=false,font=normalsize,labelfont=sf,textfont=sf]{subfig}
\usepackage{textcomp}
\usepackage{stfloats}
\usepackage{url}
\usepackage{verbatim}
\usepackage{multicol}
\usepackage{multirow}
\usepackage{threeparttable}
\usepackage{booktabs}
\usepackage{colortbl,xcolor,array}
\usepackage{xcolor}

\usepackage{algpseudocode}

\usepackage{algorithm}  
\usepackage{algorithmicx}

\usepackage{newtxmath}
\usepackage{graphicx}

\ifCLASSOPTIONcompsoc
  \usepackage[nocompress]{cite}
\else
  \usepackage{cite}
\fi
\ifCLASSINFOpdf
\else
\fi
\begin{document}

\title{Language-Guided Graph Representation Learning for Video Summarization}
\author{Wenrui Li,
        Wei Han,  
        Hengyu Man,
        Wangmeng Zuo,~\IEEEmembership{~Senior Member,~IEEE}\\
        Xiaopeng Fan,~\IEEEmembership{~Senior Member,~IEEE}
        Yonghong Tian,~\IEEEmembership{~Fellow,~IEEE}

\IEEEcompsocitemizethanks{
\IEEEcompsocthanksitem Wenrui Li, Wei Han, Hengyu Man, Wangmeng Zuo and Xiaopeng Fan are with the Department of Computer Science and Technology, Harbin Institute of Technology, Harbin 150001, China, and also with Harbin Institute of Technology Suzhou Research Institute, Suzhou 215104, China. \protect\\
E-mail: liwr@stu.hit.edu.cn; 2021111641@stu.hit.edu.cn;
manhengyu@hotmail.com; wmzuo@hit.edu.cn; fxp@hit.edu.cn
\IEEEcompsocthanksitem Yonghong Tian is with the School of AI for Science, the Shenzhen Graduate School, Peking University, Shenzhen, China, the Peng Cheng Laboratory, Shenzhen, China and also with the School of Computer Science, Peking University, Beijing, China. \protect\\
E-mail: yhtian@pku.edu.cn}
\thanks{Corresponding author: Xiaopeng Fan}}

\markboth{Journal of \LaTeX\ Class Files,~Vol.~14, No.~8, August~2015}%
{Shell \MakeLowercase{\textit{et al.}}: Bare Demo of IEEEtran.cls for Computer Society Journals}



\IEEEtitleabstractindextext{%
\begin{abstract}

With the rapid growth of video content on social media, video summarization has become a crucial task in multimedia processing. However, existing methods face challenges in capturing global dependencies in video content and accommodating multimodal user customization. Moreover, temporal proximity between video frames does not always correspond to semantic proximity. To tackle these challenges, we propose a novel \textbf{L}anguage-guided \textbf{G}raph \textbf{R}epresentation \textbf{L}earning \textbf{N}etwork (\textbf{LGRLN}) for video summarization. Specifically, we introduce a video graph generator that converts video frames into a structured graph to preserve temporal order and contextual dependencies. By constructing forward, backward and undirected graphs, the video graph generator effectively preserves the sequentiality and contextual relationships of video content. We designed an intra-graph relational reasoning module with a dual-threshold graph convolution mechanism, which distinguishes semantically relevant frames from irrelevant ones between nodes. Additionally, our proposed language-guided cross-modal embedding module generates video summaries with specific textual descriptions. We model the summary generation output as a mixture of Bernoulli distribution and solve it with the EM algorithm. Experimental results show that our method outperforms existing approaches across multiple benchmarks. Moreover, we proposed LGRLN reduces inference time and model parameters by 87.8\% and 91.7\%, respectively. Our codes and pre-trained models are available at \url{https://github.com/liwrui/LGRLN}.

\end{abstract}

}

\maketitle

\IEEEdisplaynontitleabstractindextext

% \fi
\IEEEpeerreviewmaketitle

\IEEEraisesectionheading{\section{Introduction}\label{sec:introduction}}


\IEEEPARstart{V}{ideo} contents processing is a crucial field in multimedia analysis \cite{wenrui111,Chen_02,Chen_03,zhuoyuan01,zhaorui1,zhaorui2,xiao01,xiao02}. The video summarization task aims to create concise representations of video content by selecting key frames or segments that capture the main information. This task has gained importance due to the massive daily growth in video data, especially on social media platforms \cite{wenrui222,wenrui333,wenrui444,wenrui5}. The rise of platforms like YouTube, TikTok, and Instagram has led to an explosion in short video content. As of 2023, about 900 hours of video are uploaded to YouTube every minute, amounting to approximately 1.2 million hours of content daily\footnote{https://thumbnailtest.com/stats/youtube/}. YouTube Shorts alone receive around 70 billion daily views, highlighting the significant consumption and production of short-form video content. These staggering numbers underscore the necessity for effective video summarization techniques to efficiently manage and navigate this vast amount of data \cite{wenrui6,wenrui7,bai01,Bai02,zhang01,zhang02}. Therefore, the numerous video summarization methods have been proposed in recent years \cite{intro1} \cite{intro2} \cite{intro3} \cite{intro4} \cite{intro5} \cite{intro6} \cite{intro7} \cite{intro8} \cite{intro9} \cite{intro10}. Video summarization can be broadly categorized into keyframe-based summarization and key shot-based summarization. The former selects important frames to create a static storyboard, while the latter segments the video into shots and selects informative ones to generate dynamic video skimming. In this paper, we focus on generating accurate and efficient video skimming, considering practical application scenarios \cite{intro11} \cite{intro12}.

In this work, we focus on video summarization, where a user's natural-language query guides the selection of a compact set of key frames or shots to create a temporally coherent visual summary of the video. In contrast, video captioning aims to generate free-form textual descriptions for a given video and has seen rapid progress. Li et al. \cite{lirevised1} propose a hierarchical modular network that aligns video and language semantics across entities, verbs, predicates, and sentences, enhanced by scene-graph–based learning, to advance video captioning beyond word-level supervision. Ma et al. \cite{lirevised2} introduce a style-aware, two-stage framework that incorporates caption styles and dynamically encodes video-style cues, enabling more precise and diverse video captions. Tian et al. \cite{lirevised3} propose a retrieval-inspired video captioning framework featuring a learnable token-shift module for fine-grained temporal modeling and a Refineformer for cross-attentive patch–caption integration. Ma et al. \cite{lirevised4} present a zero-shot video captioning framework that leverages learnable tokens to bridge frozen vision–language models with GPT-2 for test-time adaptation, enabling video-aware captioning without training data and yielding significant performance gains. Although related, the two tasks differ in output modality, objective, and evaluation criteria. We focus on summarization because it produces interpretable, efficient, and controllable video skims. This motivates our language-guided graph representation learning approach, which explicitly captures long-range temporal and semantic relationships while maintaining a lightweight design.

Existing video summarization methods aim to enhance the model's ability to capture global dependencies, integrate multimodal information, and detect saliency. Due to the informative and complex content of unconstrained videos (e.g., home videos or surveillance footage), Sun et al. \cite{intro2} generated montage summaries representing the video's main content by detecting salient segments. DSNet \cite{intro7} employs a dual-stream network architecture that simultaneously processes content and style information, creating more comprehensive summaries. However, traditional sequential models struggle to capture long-range dependencies in videos, which makes it difficult to produce high-quality summaries. Zhao et al. \cite{intro8} addressed this by modeling videos as graph structures and using graph convolutional networks to capture global dependencies between shots, resulting in more representative summaries. Despite these advancements, existing methods lack the flexibility to customize summaries based on user requirements. Effectively leveraging various modalities (e.g., visual, audio, and textual) is crucial for improving the quality of summaries. CLIP-It \cite{intro6} introduced a multimodal transformer guided by language to generate video summaries, allowing users to customize summaries through natural language queries and enhancing both flexibility and accuracy. QMS \cite{intro5} proposed a query-based micro-video summarization method that automatically generates short video summaries relevant to user queries. VideoXum \cite{intro9} employs a multimodal transformer to combine different modalities of video information. Different from previous graph methods, Hong et al. \cite{qigraph1} model language–visual entity relationships for navigation decisions, and Shabani et al. \cite{qigraph2} sparsify traffic graphs for forecasting. In contrast, our method represents each video frame as a node, builds forward, backward, and undirected temporal graphs, integrates language tokens through explicit edges to video nodes, and applies bi-threshold relational reasoning with a mixture-of-Bernoulli EM objective, enabling efficient and interpretable video summaries.

\begin{figure*}
	\centering
	\includegraphics[width=0.99\linewidth]{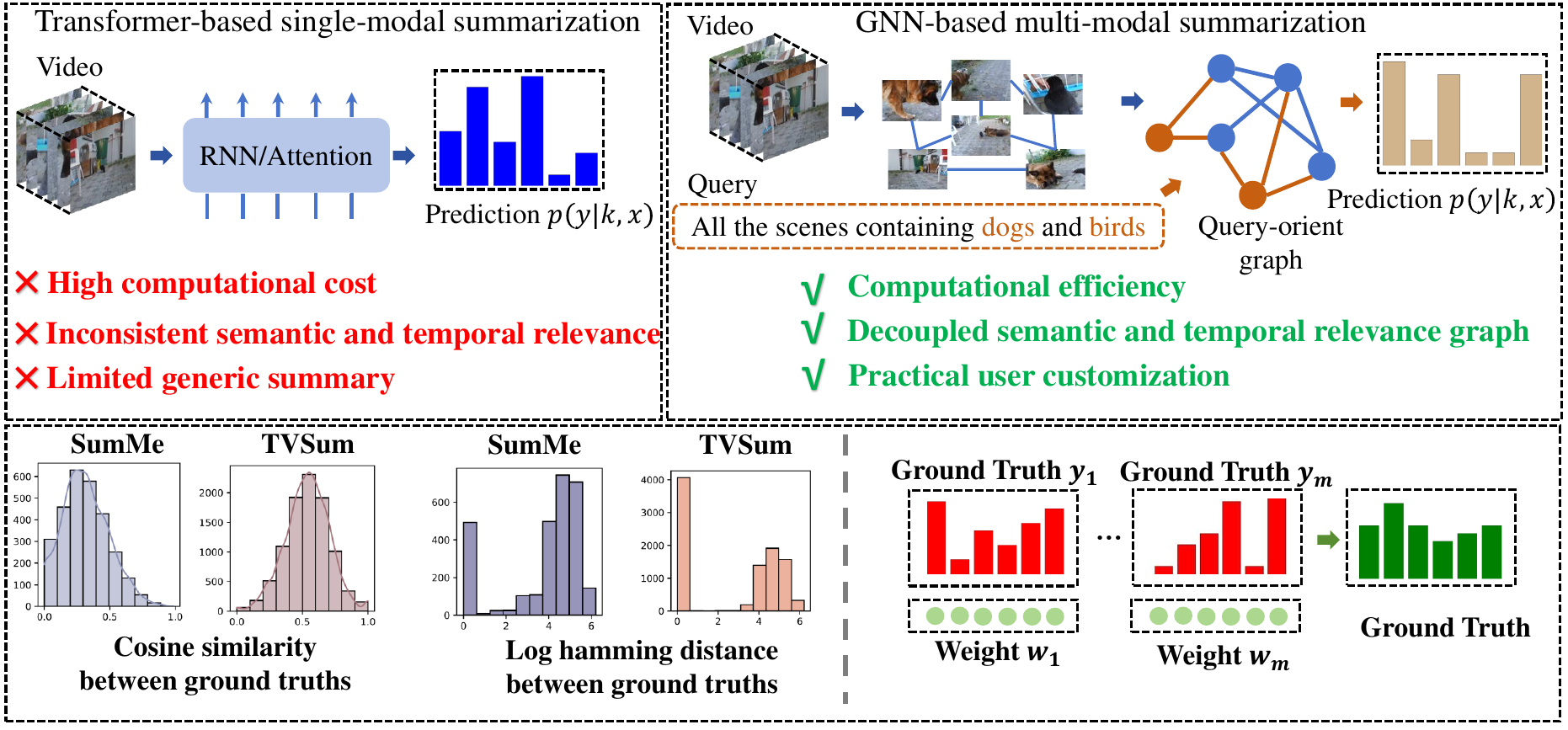}
% 	[height=8cm,width=18cm] [scale=0.78]
	\caption{The motivation of this paper. First, compared to RNN or attention-based methods, graph neural networks (GNNs) more naturally integrate multimodal information using heterogeneous graphs, while also reducing computational complexity as edge sparsity increases. Second, GNNs represent videos as graphs and conduct video summarization through node classification, providing strong interpretability. Finally, given that a video in the dataset may have multiple labels, directly averaging these labels could result in information loss, making it crucial to employ appropriate methods for processing distinct labels. To address label discrepancies in the TVSum and SumMe datasets, where cosine similarity and Hamming distance may vary between annotators, our method applies weighted averaging within an EM algorithm framework, improving label consistency and relevance in the generated summaries.}
	\label{fig:1}
\end{figure*}
A practical scenario that motivates our design is query-guided video summarization on resource-constrained devices (e.g., smartphones), where computation budget and latency strongly affect user experience. In such settings, the key technical challenge is to reduce model cost while still capturing global temporal and semantic associations and supporting language-guided personalization. Our work moves in this direction by proposing a lightweight, graph-based approach. The main challenge in video summarization is to efficiently capture global temporal relationships and semantic associations in resource-constrained environments, ensuring summary completeness and consistency. Additionally, the model must offer a high level of customization to meet diverse user needs. Unlike traditional models generating generic summaries, this model is expected to create personalized summaries tailored to individual user descriptions. It should also generalize effectively across diverse video scenarios and user inputs. Consequently, to deliver an efficient and accurate solution, the model must ensure high computational efficiency while capturing comprehensive and consistent global temporal and semantic associations in resource-limited settings. Furthermore, it should adapt flexibly to specific user needs, generating highly personalized outputs. These requirements are closely interconnected: computational efficiency ensures feasibility on resource-limited devices, capturing global associations is key to high-quality summaries, and personalization is vital for addressing user-specific needs.

In general, as illustrated at the top of Fig \ref{fig:1}, conventional methods like Transformer-based single-modal summarization face challenges like high computational costs, inconsistent semantic and temporal relevance, and limited generalization in summaries. These limitations become especially evident in scenarios requiring efficient and meaningful video summarization across diverse, large-scale datasets. To address these challenges, we propose a GNN-based multi-modal summarization framework. This method is distinguished by its computational efficiency, the decoupling of semantic and temporal relevance, and its practical user customization capabilities. By constructing a query-oriented graph that integrates multiple modalities, our model generates summaries that not only capture the intrinsic nature of video content but also align closely with specific user-defined queries, such as isolating scenes containing objects like ``dogs" and ``birds". Our method ensures that the generated summaries are semantically informative and temporally coherent, providing significant improvements over traditional methods. 

Moreover, as shown in the comparative results in the lower part of Fig. \ref{fig:1}, our method models the summary generation output as a mixture of Bernoulli distribution and solve it with the EM algorithm, outperforms traditional approaches in metrics like cosine similarity, log Hamming distance between ground truths, and in handling the one-to-many mapping challenge. The integration of these features into our model allows it to produce highly relevant, customizable video summaries while maintaining low computational cost. 

In this paper, we propose a novel \textbf{L}anguage-guided \textbf{G}raph \textbf{R}epresentation \textbf{L}earning \textbf{N}etwork (LGRLN) for video summarization. Traditional video summarization methods often struggle to accurately capture the intricate temporal and semantic relationships within video content, resulting in suboptimal outcomes. To address this challenge, our approach starts with a video graph generator that transforms video frames into a structured graph, effectively preserving both temporal order and contextual dependencies. This graph-based representation provides a solid foundation for more sophisticated summarization by maintaining the sequential nature of video content while also considering the broader context of each frame. The intra-graph relational reasoning module enhances this by employing a bi-threshold graph convolution mechanism to intelligently filter and aggregate relevant information from neighboring frames. This module effectively reduces noise by distinguishing between semantically relevant and irrelevant frames, thereby improving the precision of the generated summaries. The language-guided cross-modal embedding module represents a significant advancement in video summarization. This module integrates natural language inputs, enabling users to generate video summaries tailored to specific textual descriptions. This cross-modal capability not only enhances the adaptability of the summarization process but also offers a personalized experience by aligning video content with user-defined contexts. By integrating these three components, our architecture not only improves the accuracy and relevance of video summaries but also sets a new standard in cross-modal summarization, dynamically aligning video content with multi-modal inputs and significantly enriching the summarization process. On the SumMe dataset, LGRLN achieves an F1 score of 54.7, which represents an improvement of approximately 18.8\% over VideoSAGE. Similarly, in the TVSum dataset, it reaches an F1 score of 58.3, representing a 13.2\% improvement over SumGraph. Despite achieving high performance, LGRLN maintains efficiency with only 3 million parameters, which is a dramatic reduction of more than 99\% compared to models like V2Xum-LLaMA-7B, which has 7 billion parameters, and V2Xum-LLaMA-13B, which has 13 billion parameters. The main contributions could be summarized as follows:
\begin{itemize}
\item This paper introduces a novel Language-Guided Representation Learning Network (LGRLN) for video summarization. This framework directly integrates natural language input into the video summarization process, allowing users to generate summaries that closely align with specific textual queries.
\item We propose an intra-graph relational reasoning module based on a dual-threshold graph convolution mechanism. This method effectively reduces noise by filtering and aggregating relevant information from neighboring frames, while dynamically updating the features of nodes and edges to address variations in significance across video frames.
\item We introduce a video graph generator that converts video frames into structured graphs, preserving the temporal order and contextual dependencies of the content. To tackle the one-to-many mapping issue in video summarization, we model the summary scheme output as a mixture of Bernoulli distribution and solve it using the EM algorithm.
\end{itemize}

The rest of the paper is organized as follows: Section \ref{relatedwork} provides a comprehensive background on Abstractive and query-oriented video summarization. Section \ref{proposedmethod} offers a detailed description of the proposed LGRLN architectures. Section \ref{experiment} presents experimental results and visualizations, demonstrating the efficacy of our model. Finally, Section \ref{conclusion} summarizes the key findings and contributions of the study.

\section{Related work}
\label{relatedwork}
\subsection{Abstractive Video Summarization}                    
Supervised video summarization methods rely on manually annotated summaries as references. The straightforward approach involves designing scoring models to evaluate the summary score of each frame or shot, selecting higher-scoring ones for the summary. Gygli et al.\cite{super_1} proposed a new method and benchmark for video summarization, evaluating each frame's visual interest through "superframe" segmentation and selecting the best frames to create an informative and engaging summary. Gong et al.\cite{super_2} introduced a supervised subset selection method using the seqDPP model, which learns to select informative and diverse sets of frames from human-created summaries, achieving high-quality video summaries. Extracting robust and efficient temporal information is crucial for enhancing video summarization performance. Zhao et al.\cite{super_3} proposed the TTH-RNN model for video summarization, addressing traditional RNNs' difficulties in handling high-dimensional video features and long sequences through a tensor-train embedding layer and a hierarchical RNN structure. Zhang et al.\cite{super_4} introduced an innovative sequence-to-sequence learning model, adding a ``review encoder" to embed the predicted summary and the original video into the same semantic space, comparing their distances to preserve the video's semantic information. Recent studies have shown the limitations of sequential models in video modeling, leading to the introduction of attention mechanisms or graph convolutional networks to consider global dependencies. Chen et al.\cite{super_5} enhanced CNNs' ability to capture global relations by aggregating features globally, projecting them into an interaction space for efficient relation reasoning, and then mapping the relation-aware features back to the original coordinate space. Zeng et al. \cite{super_6} constructed a graph of action units and their relationships, using a graph convolutional network (GCN) to capture the temporal and semantic relationships between action units, enhancing temporal action localization in videos. VideoSAGE \cite{chaves2024videosage} constructs a sparse graph to capture long-range frame relationships and formulates the summarization task as a binary node classification problem. Mrigank et al.\cite{super_7} proposed a fully convolutional sequence model for video summarization, treating video summarization as a sequence labeling problem and using methods from semantic segmentation to select important frames for summary videos.

Unsupervised video summarization methods focus on designing learning-based criteria, transforming the task into a subset selection problem. Early approaches leveraged clustering algorithms to group frames or shots into clusters, defining the center of each cluster as key-frames or key-shots. Lee et al. \cite{unsupevised1} utilized region cues like hand proximity and gaze to predict importance, proving effective without being specific to any user or object. Mundur et al. \cite{unsupevised2} presented a fully automatic video summarization technique using Delaunay Triangulation for clustering frames, producing high-quality summaries with fewer frames and less redundancy than other methods. To address the insufficient capture of fine-grained contextual scene interactions and motion information by clustering algorithms, dictionary learning converts the problem into a sparse optimization problem. Elhamifar et al. \cite{unsupevised3} introduced a method to find representative data points that describe an entire dataset through a sparse multiple measurement vector problem, selecting representatives via convex optimization without assuming low-rank or cluster-centered data. SumGraph \cite{park2020sumgraph} refines frame-level semantic relationships via graph convolutions, enabling effective keyframe selection in both supervised and unsupervised settings. RSGN \cite{intro8} integrates LSTM for intra-shot frame dependencies and GCN for inter-shot relationships, enabling unsupervised learning through reconstruction loss to capture both local and global video context effectively. DSAVS \cite{DSAVS} selects semantically rich summaries by aligning video and text representations, enhanced with self-attention to capture long-range temporal dependencies.

However, traditional video summarization methods struggle with long-duration videos due to weak global dependencies, inadequate modeling of local-global relationships, and the diversity of video summaries. Our method leverages a video graph generator to transform video frames into structured graphs, thereby preserving temporal order and contextual dependencies more effectively. Our dual-threshold graph convolution mechanism effectively reduces noise, dynamically updates node and edge features, and ensures the proper modeling of frame importance. Additionally, we employ a hybrid Bernoulli distribution to model the summarization process and optimize it using the EM algorithm, effectively addressing the one-to-many mapping problem in video summarization.

\subsection{Query-oriented Video Summarization}
Current video summarization methods generating generic summaries are inadequate for meeting the personalized needs of different users. Consequently, generating summaries focusing on specific parts based on users' natural language descriptions has become a crucial challenge. Sharghi et al. \cite{query1} proposed a probabilistic model to select key shots from long videos based on their relevance. Sharghi et al. \cite{query2} presented a memory network-based model for query-focused summaries, addressing user subjectivity and evaluation challenges by emphasizing semantic information over visual features. Kanehira et al. \cite{query3} introduced viewpoint-specific video summarization, leveraging semantic similarities and differences among video groups to optimize diversity, representativeness, and discriminativeness. Jia et al. \cite{intro5} introduced a query-oriented micro-video summarization model employing an encoder-decoder transformer to handle diverse entities and complex scenes. Narasimhan et al. \cite{intro6} proposed a multimodal transformer framework for both generic and query-focused video summarization, utilizing language models to guide importance scoring.

Traditional query-focused video summarization methods struggle to accurately capture query relevance and lack explicit structured modeling. Our approach incorporates a language-guided cross-modal embedding module, seamlessly integrating natural language inputs to generate video summaries closely aligned with specific textual queries. Utilizing a GNN-based framework, our method effectively fuses multimodal information, ensuring high semantic consistency between generated summaries and user queries.

\subsection{Large Language Models in Video Understanding}
In recent years, video understanding methods based on Large Language Models (LLMs) have demonstrated outstanding performance in video signal processing, thanks to their powerful reasoning capabilities. For example, Videopoet \cite{LLM1} can synthesize high-quality videos from multiple conditional signals and match them with corresponding audio, showcasing the potential of LLMs in video generation. Unlike traditional methods that encode images and videos into different feature spaces and input them separately into LLMs, Video-LLaVA \cite{LLM2} innovatively unifies visual representations into a single language feature space, fundamentally addressing the issue of projection mismatch caused by the lack of a unified tokenization between images and videos. This significantly enhances the efficiency of multimodal understanding. VideoLLM \cite{LLM3}, inspired by the exceptional causal reasoning capabilities of LLMs, transfers these reasoning abilities to video sequence understanding tasks. By introducing the Modality Encoder and Semantic Translator, it converts video frames and other multimodal inputs (e.g., text) into a unified token sequence, overcoming the limitations of task specialization in existing video understanding models and enabling more flexible multitask handling. Furthermore, Videochat \cite{LLM4} integrates LLMs with video foundation models via a learnable neural interface, empowering the system with powerful video content analysis and dialogue capabilities, driving video understanding towards more interactive and human-like experiences. Finsta \cite{LLM5} introduces an innovative fine-grained spatiotemporal alignment approach that leverages scene graph modeling and alignment techniques, enhancing the cross-modal representation ability of video language models and improving the model’s understanding of complex video content.

In the field of video summarization, LLM-based methods have also garnered widespread attention \cite{VLLM1,VLLM2,VLLM3,VLLM4}. V2Xum-LLM \cite{vlm_videoxum} proposed an innovative framework that unifies various video summarization tasks within the LLM text decoder, significantly improving performance and flexibility by leveraging multimodal support and unified task modeling. M3Sum \cite{vlm_m3sum} simplifies the cross-modal alignment issue with a parameter-free alignment mechanism and leverages LLMs' strength in long-text understanding to effectively enhance the quality of video summarization.

Despite significant performance breakthroughs achieved by LLM-based video understanding methods, they still face high training costs and inference times, which pose a major challenge to their deployment in resource-constrained environments. As a result, developing high-performance, low-power video summarization models has become a key research direction. This study presents an efficient and lightweight LLM-driven video summarization method that reduces computational resource consumption while maintaining inference effectiveness, offering a new solution for video processing in real-world applications. Experiments on standard benchmarks validate the proposed method’s efficiency and effectiveness.
% \subsection{Unsupervised Video Summarization}
 % Lu et al. \cite{unsupevised4} used locality-constrained linear coding (LCC) to transform raw local features into sparse representations and weighted their importance to capture dominant content and discriminative details. Additionally, adversarial learning has shown excellent performance by classifying between indistinguishable and ground-truth summaries. Yuan et al. \cite{unsupevised5} proposed an unsupervised video summarization model using a cycle-consistent adversarial LSTM architecture to maximize information preservation and compactness. Fu et al. \cite{unsupevised6} introduced a GAN-based video summarization approach, featuring an attention-aware Ptr-Net generator to create summarization points and a 3D discriminator to distinguish between ground-truth and generated fragments.

\section{The Proposed Approach} \label{proposedmethod}
% We introduce a graph bi-threshold network architecture that treats video summarization as a graph node classification task. The architecture primarily comprises four components: a video graph generator, a graph bi-threshold module, a text embedding module, and a bi-threshold cross entropy loss function. The video graph generator models the input video as three types of graphs: forward, backward, and undirected, based on the time interval between frames and the direction of the video stream; The graph bi-threshold module, fed by the video graph, computes the importance score for each node; The text embedding module treats both natural language sequences and video sequences as interconnected graphs, integrating external natural language instructions into the video sequence to produce personalized summaries; The bi-threshold loss function accommodates varying human annotation preferences. It conceptualizes the model output in terms of a mixed Bernoulli distribution, thereby aligning the model's output with specific annotated preferences rather than all possible annotations, which could lead to the convergence objective straying from a reasonable range. The overall structure of the model is shown in Figure 1.

\begin{figure*}
	\centering
	\includegraphics[width=0.99\linewidth]{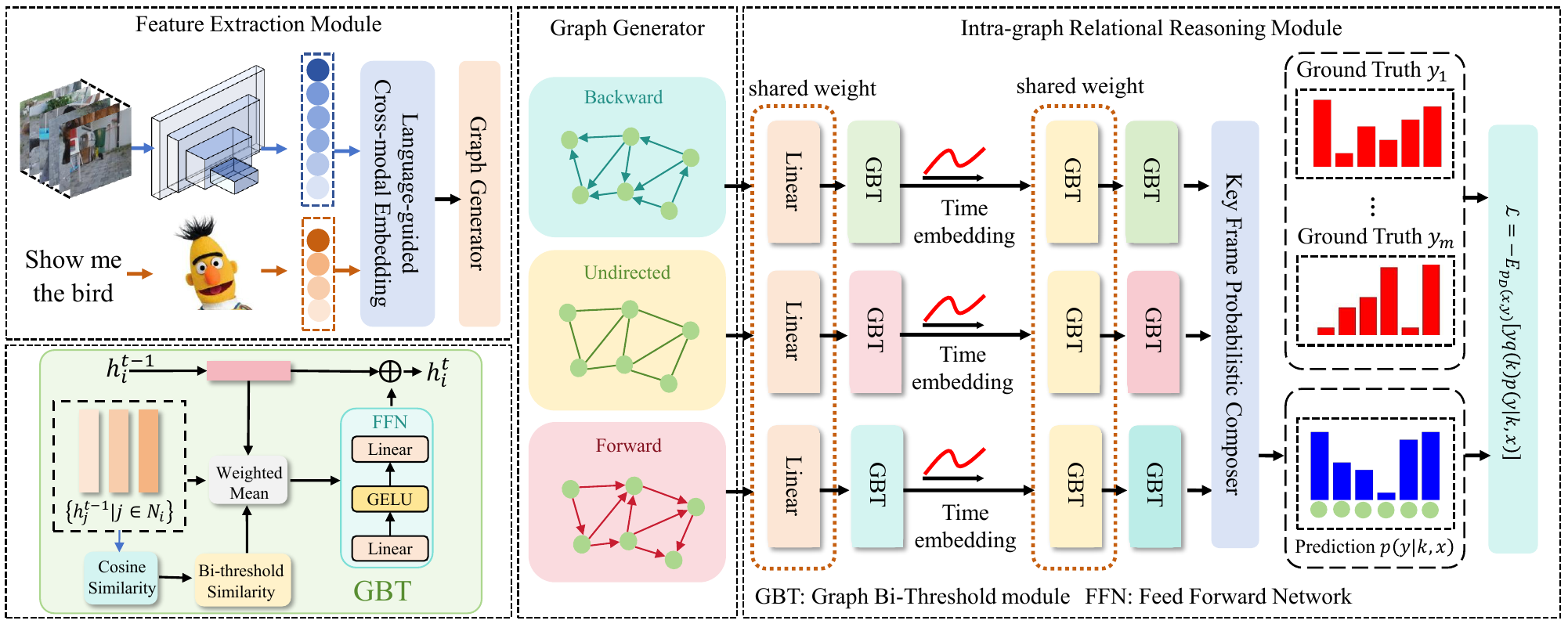}
% 	[height=8cm,width=18cm] [scale=0.78]
	\caption{The overall architectures of proposed method. Input features from video and text modalities are fused using the Language-Guided Cross-Modal Embedding module. The fused input is then transformed into three graphs by the Graph Generator, which are subsequently fed into the three branches of the intra-graph relational reasoning Module. Outputs from all branches are combined by addition to generate the model's prediction. Finally, BCE loss is used to compute the overall loss between the prediction and the ground truths.}
	\label{fig:2}
\end{figure*}

The proposed architecture consists of four primary components: a video graph generator, a intra-graph relational reasoning module, a language-guided cross-modal embedding module, and a bi-threshold cross-entropy loss function. The video graph generator models the input video as three types of graphs: forward, backward, and undirected, based on the time intervals between frames and the direction of the video stream, to capture information from different directions of the video stream. The intra-graph relational reasoning module calculates the importance score for each node and designs aggregated weights combined with variational inference, combining prior knowledge to reduce model sensitivity and enhance robustness. The language-guided cross-modal embedding module considers both natural language sequences and video sequences as interconnected graphs, integrating external natural language instructions into the video sequence to generate personalized summaries. This information fusion method based on heterogeneous graphs can explicitly design information aggregation patterns through edge construction rules, thereby improving model interpretability and controllability. The bi-threshold loss function adapts to varying human annotation preferences. It models the output as a mixed Bernoulli distribution, aligning the model's output with specific annotated preferences rather than all possible annotations, thereby reducing the loss of label information. Tab. \ref{tab1} demonstrated the notations and descriptions in detail.
\begin{table}
\centering
		\renewcommand\arraystretch{1.4}
\caption{Key Notations and Descriptions}
\label{tab1}
\begin{tabular}{|c|c|}
\hline
\textbf{Notation} & \textbf{Description} \\ \hline
$G_f(V, E_f)$ & Forward video graph (nodes $V$, edges $E_f$) \\ \hline
$G_b(V, E_b)$ & Backward video graph (edges $E_b$) \\ \hline
$G_u(V, E_u)$ & Undirected video graph (edges $E_u$) \\ \hline
$\boldsymbol{x}_i \in \mathbb{R}^{D_x}$ & Frame $i$ feature vector ($D_x$: dimension) \\ \hline
$\mathcal{G}(V, E)$ & General graph (nodes $V$, edges $E$) \\ \hline
$\boldsymbol{h}_i^{(t)}$ & Node $i$ feature at iteration $t$ \\ \hline
$\boldsymbol{\alpha}_i^j$ & Aggregation weight (cosine similarity) \\ \hline
$\boldsymbol{\tau}_e$ & Time embedding for temporal order \\ \hline
$L_{BCE}$ & Bi-threshold cross-entropy loss \\ \hline
$p(\boldsymbol{y}_i = 1 | \boldsymbol{z}_i)$ & Probability node $i$ is a keyframe \\ \hline
\end{tabular}
\end{table}

\subsection{Task Description}
The video summarization task is to find a mapping $ f: \mathbb{R}^{\left| V \right | } \rightarrow \mathbb{R}^{\left| S \right | } $, taking the video sequence $ V $ as input and subset $ S $ of $ V $ as output, and ensure that the subset $ S $ can contain the main semantic information of $ V $. In the model-based approach,  video $ V $ will be firstly extracted frame by frame to obtain a feature sequence $ \left \{ x_i \right \}_{i \in V}  $, and then a neural network is trained on the dataset to fit the mapping $ f $. In the training set, each video sequence has $ k $ ground-truth summaries $ \left \{ y_i \right \}_{0 \le i < k} $ generated by different annotators. Each annotation sequence $ y $ is a binary sequence with a value of 0 or 1, where whether the $ j $-th element is 1 represents whether the $ j $-th frame belongs to the digest subset $ S $. Compared to the traditional video summarization tasks, the video summarization task guided by natural language adds text information $ t $ as additional input to the mapping $ f $ to control the mapping behavior. This requires the mapping $ f $ to consider more complex multimodal information.

\subsection{Video Graph Generator}
Consistent with VideoSAGE \cite{chaves2024videosage}, we represent the input video sequence as three graphs: a forward graph $ G^f(V,E^f) $, a backward graph $ G^b(V,E^b) $, and an undirected graph $ G^u(V,E^u) $. The reason for modeling video sequences as three types of graphs is to incorporate additional temporal ordering. If only an undirected graph is used, for any node, both the previous and next frames are its neighbors. The undirected graph is not clear about the chronological order of these two frames, so it is necessary to add a unidirectional flow graph of information, such as a forward graph. The reason for adding a backward graph is to bring information about the video after each frame, which helps the node locate its role in the entire video sequence.

Each video frame $ i \in V $ is considered a node in the graph, with the representation vector $ \boldsymbol{x}_i \in \mathbb{R}^{D_{x}} $ representing the features extracted from that frame using GoogLeNet \cite{szegedy2015going}. If the time interval between two frames is less than the threshold $ \boldsymbol{\tau} $, an edge connects them. In the undirected graph $ G^u $, the edge is undirected. In the forward graph $ G^f $, the edge is directed from the earlier frame to the later frame, while in the backward graph $ G^b $, it flows in the opposite direction. After generating the graph representation of the video sequence, we use the intra-graph relational reasoning Module to extract relevant information from the graph and generate summarization scores.

\subsection{Intra-graph Relational Reasoning Module}
\subsubsection{Graph Bi-Threshold Module}

Due to the excellent performance of graph convolutional networks in graph node classification tasks, we employ graph convolution \cite{zhang2019graph} for node classification, extracting sufficient features from the three graphs generated by the Video Graph Generator to generate summarization scores. The graph convolutional neural network, consists of a permutation-invariant aggregation function $ g: \mathbb{R}^{(\left | N_{i} \right | +1) \times D_{in}} \rightarrow \mathbb{R}^{D_{in}}$ and a permutation equivariant iterative function $ f: \mathbb{R}^{2 \times D_{in}} \rightarrow \mathbb{R}^{D_{out}}   $, where $ N_i $ is the set of neighbors adjacent to node $i$, $ D_{in} $ and $ D_{out} $ are the input and output dimensions of each node in the graph network. 

The aggregation function $ g $ summarizes the features of the neighboring nodes $ \left \{ \boldsymbol{h}_{j}^{t-1} | j \in \boldsymbol{N}_{i} \right \}  $ in the graph, capturing the overall impact of information propagation from neighboring nodes to the target node. The iteration function $ f $ integrates the summarized information with the current node information $ \boldsymbol{h}_{i}^{t-1} \in \mathbb{R}^{D{in}} $ to generate the updated node information $ \boldsymbol{h}_{i}^{t} \in \mathbb{R}^{D{out}} $. The process of graph convolution can be represented by :
\begin{equation}
    \boldsymbol{h}_{i}^{t} = f(\boldsymbol{h}_{i}^{t-1}, g(\boldsymbol{h}_{i}^{t-1}, \boldsymbol{h}_{j}^{t-1} | j \in \boldsymbol{N}_{i})),
\end{equation}
where $f(\cdot)$ is the iterative function, $g(\cdot)$ is the aggregation function, $\boldsymbol{h}_i^t$ represents the iterative result of the $i$-th node in the $t$-th layer, and $\boldsymbol{N}_i$ represents the set of neighboring nodes of $i$. 

The aggregation functions in classic graph convolutional networks (GCN \cite{kipf2016semi}, GAT \cite{velivckovic2017graph}, SAGE \cite{hamilton2017inductive}, etc.) combine all neighbor information. This approach works well when the graph structure is clearly defined. However, if the graph structure contains errors, such as edges between nodes that should not be neighbors, noisy information can be erroneously combined with the target node. For example, in videos, camera jumps, slow motion, close-ups, and other operations may cause adjacent frames to be semantically non-adjacent, even if the time interval between these frames is within the threshold. Due to special operations such as shot jumps and close ups in videos, graphs generated solely based on time information in Video Graph Generator will connect many frames that are not semantically neighbors. This will result in traditional graph networks incorporating interference information from erroneous neighbors when processing nodes. Therefore, we need to develop a new graph network structure that eliminates interference information from erroneous neighbors based on the semantic information of nodes.

Variational inference can be used to analyze which nodes are semantically true neighbors. For any node $ h_i $, the event that $ h_j $ is a neighbor of $ h_i $ is defined as $ z $, and the posterior probability of $ z $ is $ 
p(z| h_i, h_j) $. Similar to the common practice of graph representation learning~\cite{graph_randomwalk}, we define the posterior probability as $ softmax(cosine(h_i, h_j)) $, and the probability modeled by our model is $ q(z) $. However, in the early stages of training, graph networks have limited representation capabilities, and at this time, $ h_i $ and $ h_j $ may not be extracted accurately. In order to increase the robustness of the model, we need to provide strong priors for the distribution of $ q(z) $ to reduce model sensitivity. The prior we propose is that $ q(z) $ is a ternary distribution with values of $ 0 $, $ \alpha_1 $, or $ \alpha_2 $, where $ \alpha_1 $ and $ \alpha_2 $ are constants. This ensures that even during network training, updates to the representation vectors of graph nodes will not significantly cause changes in $ q (z) $, resulting in stable convergence. We use variational inference to obtain the actual value of $ q (z) $ in the network, specifically:

\begin{equation}
    \max_{q(z)} \sum_zq(z)log(\frac{p(z, x_i, x_j)}{q(z)}).
\end{equation}

By using the solution result of q (z) as the aggregation weight, we propose the aggregation function $ g $, considers the cosine similarity between the representation of node $ \boldsymbol{h}_{i}^{t-1} $ and the representations of neighboring nodes $ \left \{ \boldsymbol{h}_{j}^{t-1} | j \in \boldsymbol{N}_{i} \right \} $ when aggregating the neighbor information of node $ i $. Neighbors with low similarity are considered semantically non-adjacent, and their aggregation weight is reset to 0. The aggregation weight of close neighbors with high similarity is reset to $ {\alpha}_{1} $, while the weight for the remaining neighbors is set to $ {\alpha}_{2} $. The aggregation function $ g $ is designed as follows:
\begin{equation}
\begin{aligned}
    &\boldsymbol{m}_i^t = g(\boldsymbol{h}_{i}^{t-1}, \boldsymbol{h}_j^{t-1} | j \in \boldsymbol{N}_i) =  {\textstyle \sum_{j \in \boldsymbol{N}_{i}}} {\alpha}_{ij}^{t}\boldsymbol{h}_{j}^{t-1}, \\
    &{\alpha}_{ij}^t = \left\{\begin{matrix}
  0,&\frac{\boldsymbol{h}_{i}^{t-1}\cdot \boldsymbol{h}_{j}^{t-1}}{\left \| \boldsymbol{h}_{i}^{t-1} \right \|\left \| \boldsymbol{h}_{j}^{t-1} \right \| } < {\tau}_{1},  \\
  {\alpha}_{1},&\frac{\boldsymbol{h}_{i}^{t-1}\cdot \boldsymbol{h}_{j}^{t-1}}{\left \| \boldsymbol{h}_{i}^{t-1} \right \|\left \| \boldsymbol{h}_{j}^{t-1} \right \| } > {\tau}_{2}, \\
  {\alpha}_{2},& otherwise,
\end{matrix}\right. 
\end{aligned}
\end{equation}
where $\boldsymbol{m}_i^t$ represents the aggregation result of the $i$-th node neighbor in the $t$-th layer, $\alpha_{ij}^t$ represents the attention of the $i$-th node to the $j$-th node in the $t$-th layer, $\left \| \cdot  \right \| $ represents the 2-norm, $\tau$ is the threshold. 

The iterative function $ f (\cdot) $ calculates the change in node information $ 
{h}_{i}^{t-1} $ based on the aggregation result and normalizes it to avoid gradient explosion. This process can be expressed as:
\begin{equation}
\begin{aligned}
    f({\boldsymbol{h}}_{i}^{t-1}, \boldsymbol{m}_i^t) = GN({\boldsymbol{h}}_{i}^{t-1} + {\boldsymbol{W}}_{2}\sigma({\boldsymbol{W}}_{1}\boldsymbol{m}_i^t + {b}_{1}) + {b}_{2}),
\end{aligned}
\end{equation}
where the nonlinear activation function $ \sigma (\cdot) $ uses GELU \cite{hendrycks2016gaussian} to capture complex nonlinear information, the normalization function $ GN (\cdot) $ represents Graph Normalization \cite{cai2021graphnorm} to accelerate the training speed, and $\boldsymbol{W_i^{t}}$ and $b$ represent the weight metric and bias term in the $i$-th MLP layer.

The output of Graph Bi-Threshold (GBT) module $ \boldsymbol{z}_i $ from the final layer is designated as the score for determining whether node $ i $ is a summary frame. When this score is passed through the sigmoid function, it becomes the probability that node $i$ is a summary frame. This process can be expressed as:
\begin{equation}
    p(\boldsymbol{y}_i=1|\boldsymbol{z}_i) = Sigmoid(\boldsymbol{z}_i),
\end{equation}
where $\boldsymbol{y}_i$ is the binary classification result of the $i$-th node, and $\boldsymbol{z}_i$ is the GBT output of the $i$-th node.

\subsubsection{Time Embedding}
When modeling a video as a graph, simply considering two frames with time intervals less than a threshold as neighbors can result in the loss of certain temporal features that represent the sequence order. For example, if three frames are arranged in chronological order as $ \boldsymbol{x}_1 $, $ \boldsymbol{x}_2 $, and $ \boldsymbol{x}_3 $, and the time interval between $ \boldsymbol{x}_1 $ and $ \boldsymbol{x}_3 $ is less than the threshold $ \tau $, then both $ \boldsymbol{x}_2 $ and $ \boldsymbol{x}_3 $ will be considered neighbors of $ \boldsymbol{x}_1 $. From the perspective of $ \boldsymbol{x}_1 $, these two frames appear identical, and it is not apparent that $ \boldsymbol{x}_2 $ is actually closer to $ \boldsymbol{x}_1 $. Therefore, it is necessary to incorporate the temporal information of each frame into the input of the graph neural network.

In certain hidden layers of the network, time embedding is incorporated to modify the hidden layer representation:
\begin{equation}
    \begin{aligned}
        \boldsymbol{h}_{i}^{t} \leftarrow {\boldsymbol{h}_{i}^{t} + \boldsymbol{\tau}_{i}^e},
    \end{aligned}
\end{equation}
where $\boldsymbol{\tau}_i^e$ is the time embedding of the $i$-th node, that allows the model to dynamically update the time representation during training.

\subsection{Language-Guided Cross-Modal Embedding Module}
Traditional information fusion methods such as cross attention mechanism use fully learnable weights for fusion. This behavior of entrusting all behavior to autonomous learning by the network can lead to loss of controllability. Network behavior is completely driven by data, and it is difficult to ensure that the network learns sufficiently robust strategies in the case of limited data. Therefore, we integrate text nodes and video nodes into heterogeneous graphs to achieve multimodal information fusion. By manually constructing edges of heterogeneous graphs, we can explicitly control the behavior of information fusion, thereby ensuring controllable network behavior.

To enable video summarization guided by natural language, we incorporate the user's natural language as an integral part of the model's input. Since video sequences are already represented as distinct nodes within a graph, it is logical to treat natural language sequences as nodes as well. Consequently, connections are established between nodes representing video and natural language, forming a heterogeneous graph $ G(\boldsymbol{x}_v, \boldsymbol{x}_s, E) $. In this graph, video nodes $ \boldsymbol{x}_v $ are defined as before, while text nodes $ \boldsymbol{x}_s $ represent tokens from the natural language sequence $ s $, with node representation vectors $ \boldsymbol{x}_s \in X_s $ being the feature vectors extracted by BERT \cite{devlin2018bert}. To fully integrate linguistic information into the video sequence, each token node is connected to all video nodes via directed edges.

The primary objective of this module is to incorporate natural language sequence information into the video sequence effectively. Since the integrated video sequence will be represented as a graph with edges between adjacent frames and then processed by the GBT. Because BERT already accounts for the interrelations between natural language tokens, this layer would disregard all edges except those from natural language tokens to video nodes to minimize parameters, focusing solely on integrating natural language information into the video sequence. Due to the complexity of text features, we utilize a graph attention mechanism instead of bi-thresholds as weights for the aggregation function $ g $, which can be defined as follows:

\begin{equation}
    \begin{aligned}
        g(\boldsymbol{h}_i^{t-1}, \boldsymbol{h}_j^{t-1} | j \in N_i) = {\textstyle \sum_{j \in N_i} {\alpha_{ij}^{t}(\boldsymbol{W}_1^th_j^{t-1}+b_1^t)}},  \\
        \alpha_{ij}^t = \frac{exp((\boldsymbol{W}_2^th_i^{t-1}+b_2^t)'(\boldsymbol{W}_3^th_j^{t-1}+b_3^t))}{ {\textstyle \sum_{j \in N_i} {exp((\boldsymbol{W}_2^th_i^{t-1}+b_2^t)'(\boldsymbol{W}_3^th_j^{t-1}+b_3^t)}} },
    \end{aligned}
\end{equation}
where $\alpha$ is the attention that integrates textual information into video nodes, $\boldsymbol{W}_i^{t}$ and $b_i^{t}$ are weight metric and bias term of the $i$-th node, respectively.

The iterative function $ f(\cdot) $ retains the same structure as GBT, employing two layers of MLP and a GELU activation function to integrate complex information from the text.

\subsection{Biased Cross Entropy Loss}
For the same video, different annotators may produce varied summaries based on their preferences. To leverage multiple labels from different annotators for the same video, the conventional approach is to compute the arithmetic mean of these labels as the summary score, and then use cross-entropy or MSE loss to calculate the error in the model's output. However, these methods discard the preference information embedded in these labels. For example, the average of [1,0] and [0,1] is [0.5, 0.5], which represents the highest entropy (uncertainty). In light of this, we introduce the use of BCE loss for video summarization, reevaluating the significance of each label by employing a weighted mean instead of an arithmetic mean to achieve a convergence target that incorporates preferences.

Given the possibility of multiple summarization schemes for the same video, we model the output using a mixed Bernoulli distribution and derive the preference cross-entropy loss (BCE) as the loss function for LGRLN video summarization training. If the video feature is $x$ and the label is $y$, this results from a specific summarization strategy. The same video may have multiple summarization strategies $k$, such as extracting the first or last few frames of each scene with the same shot. However, when the strategy is well-defined, there will be only one summarization scheme per video under that strategy. Therefore, the probability of modeling video summarization is $ p(y|x) =  {\textstyle \sum_{k} {p(k)p(y|x,k)}} $. Under this model,  the log-likelihood function of the dataset $ D $ is $ E_{p_{D}(y,x)}\left [ log(p(y|x)) \right ] $, and $ log(p(y|x)) $ can be decomposed into $ {\textstyle \sum_{k} {q(k)log(\frac{p(y,k|x)}{q(k)} \frac{q(k)}{p(k|y,x)})}} $, which allows us to use the EM algorithm \cite{mclachlan2007algorithm} to solve the maximum likelihood problem of this mixed distribution. 

The closest posterior probability $ q(k) $ is obtained in step E, which can be written as:
\begin{equation}
    \min_{q(k)} {{E}_{p_{D}(y|x)}\left [  {\textstyle \sum_{k} {q(k)log(\frac{q(k)}{p(k|y,x)} )}}  \right ],} 
\end{equation}
where $\sum_{k} {q(k)log(\frac{q(k)}{p(k|y,x)}}) $ can be decomposed into:
\begin{equation}
\begin{aligned}
    \sum_{k} {q(k)log(q(k))}-\sum_{k} {q(k)log(p(y|k,x))}\\
    -\sum_{k} {q(k)log(p(k|x))}+log(p(y|x)),   
\end{aligned}
\end{equation}
where $q$ can be any distribution used to assist in training the model distribution $p$.

To simplify the model, assume $q(k)$ is a binary function, which makes its entropy $ -\sum_{k} {q(k)log(p(k|x))} $ a constant. Meanwhile, due to the lack of a clear arrangement order for $k$, the prior distribution $p(k|x)$ is uniform, making $ log(p(k|x)) $ a constant. Additionally, in the objective function for $q(k)$, $p(y|x)$ is also a constant. Therefore, the above minimization problem is equivalent to maximizing $ \sum_{k} {q(k)log(p(y|k,x))} $, which has an analytical solution:

\begin{equation} \label{q}
\begin{aligned}
    q(k) = &\left\{\begin{matrix}
  b,& k \in S(x,y), \\
  a,& otherwise,
\end{matrix}\right. \\
&s.t. \\
0 \le &a < b < 1, \\
\sum_{k} &q(k) = 1, \\
\forall k_1 \in S, k_2 \in \bar{S}, &log(y|k_1, x) > log(y|k_2, x),
\end{aligned}
\end{equation}
where $a$ and $b$ are hyperparameters, and the larger the difference between them, the greater the entropy of $q$. $S$ is the subset closest to the model output $x$ in the ground truth $y$, and $\bar{S}$ is the complement of all ground truths to $S$.

The model parameters only need to be optimized in step M. Step M involves solving the maximization problem, which can be defined as:
\begin{equation} \label{M}
    \max_{p(y|k,x)} {E_{p_{D}(y,x)}\left [  {\textstyle \sum_{k} { q(k)log(\frac{p(y,k|x)}{q(k)} ) }}  \right ], }
\end{equation}
where $P_D$ is the distribution in the dataset, and $k$ is the latent variable, representing which strategy to use for video summarization. Among them, Eq. (\ref{M}) can also be decomposed into:
\begin{equation}
    \begin{aligned}
        \sum_{y} {\sum_{k} {p_{D}(y,x)q(k)log(p(y|k,x)) }}+\\
        \sum_{y} {p_{D}(y,x)\sum_{k} {q(k)log(p(k|x)) }}-\\
        \sum_{y} {p_{D}(y,x)\sum_{k} {q(k)log(q(k)) }}.
    \end{aligned}
\end{equation}

As in the analysis of step E, $ \sum_{k} {q(k)log(p(k|x))} $ and $ \sum_{k} {q(k)log(q(k)) } $ are constants variables, so only $ \sum_{y} {\sum_{k} {p_{D}(y,x)q(k)log(p(y|k,x)) }} $ needs to be optimized. To further simplify the model, it is required that the model only needs to learn the summary scheme under one strategy, meaning it only needs to model one corresponding output. In this case, the maximization problem in step M is simplified to $ \sum_{y,x} {p_{D}(y,x)q(k)log(p(y|k,x)) } $. Since whether a frame is a keyframe $y$ is a binary variable, the above equation is equivalent to:
\begin{equation} \label{bce}
\begin{aligned}
    \sum_{y,x} {p_{D}(y,x)q(k)(ylog(p(+|k,x)) 
    + (1 - y)log(p(-|k,x))) },
\end{aligned}
\end{equation}
where $p(+|k,x)$ is the probability that the model considers frame $x$ a keyframe in policy $k$, and $p(-|k,x)$ is the probability that the model considers $x$ a non keyframe, $y$ is the ground truth, and $q (k)$ comes from Eq. (\ref{q}), $p$ is the predicted of the model. The difference between the above equation and standard cross-entropy is that the label is weighted by $q(k)$, which is referred to as biased cross-entropy.

This loss function enables the model output to approximate a subregion of the reasonable interval, rather than converging to the mean of the interval, which may lie outside the reasonable range.

% \begin{figure}
%     \centering
%     \includegraphics[width=0.5\linewidth]{image.png}
%     \caption{Convergence direction with BCE loss comparing to CE.}
%     \label{fig:enter-label}
% \end{figure} 

\section{Experiments}\label{experiment}

\subsection{Experimental Settings}

\subsubsection{Datasets}
Our experiments are primarily conducted on two benchmark datasets: SumMe \cite{gygli2014creating} and TVSum \cite{song2015tvsum}. SumMe consists of 25 videos from diverse users and scenarios (e.g., cooking, racing), with each video annotated by 15-18 users to indicate whether each frame is a summary frame. TVSum comprises 50 videos from YouTube, each annotated by 20 users. Additionally, we employed the VideoXum \cite{intro9} dataset and the QFVS \cite{akhare2022query} dataset in our analysis. The VideoXum dataset, which consists of over 10,000 videos, serves as a valuable resource for assessing the model's ability to extract insights from large-scale data.

\subsubsection{Evaluation Protocol}
In line with established practices, a 5-fold cross-validation method is applied to the SumMe, TVSum, and VideoXum datasets. F1 scores are calculated on 4 splits for the QFVS dataset. The datasets are randomly divided into five equal parts, with 20\% designated as the test set and the remaining 80\% used as the training set for the experiments. The final experimental result is obtained by averaging the outcomes of the five individual experiments. For VideoXum, we employ a distinct approach, using a partitioned dataset with a separate training set for model training, a test set for monitoring training progress, and a validation set for evaluating the final results. For evaluation criteria, we follow industry standards, utilizing the F1 score of the summary results, along with the $ \tau $ and $ \rho $ scores, to assess our model's performance. The F1 score is calculated as follows:
\begin{equation}
    \begin{aligned}
        F_1=\frac{2PR}{P+R},
    \end{aligned}
\end{equation}
where $P$ is the accuracy calculated using $ |S\cap\widehat{S}|/|\widehat{S}| $, $R$ is the recall calculated using $ |S\cap\widehat{S}|/|S| $, $S$ is the ground truth and $\widehat{S}$ is the model prediction value. Additionally, $\tau$ and $\rho$ represent the Kendall and Spearman correlation coefficients, respectively, which quantify the correlation between the model's output scores and the ground truth labels.
\begin{algorithm}[t]
\caption{Training framework for the proposed approach}
\label{algo1}
\begin{algorithmic}[1]

\Require Dataset $X$, where in any item $(x, t, y)$ of the dataset, $x$ is a video, $t$ is the natural language guidance of the video summary, and $y$ is the $m$ ground truths of the manually provided video summary.

\Ensure Updated network parameters.

\State Extract features from the video and text: $(v,t)$ in the dataset as $(x_v, x_t)$.
\For{each epoch}
    \For{$(x_t, x_v, y)$ in $X$}
        \State $ V \gets LGCME(x_t, x_v)$ \Comment{Language Guided Cross Modal Embedding Module}
        \State $ G^f, G^b, G^u \gets VGG(V)$ \Comment{Video Graph Generator}
        \State $ \hat{y} \gets IGRR(G^f, G^b, G^u) $ \Comment{Intra-graph Relational Reasoning Module}
        \State $ L \gets BCE(\hat{y}, y) $
        \State Optimize model parameters using optimizers such as AdamW.
    \EndFor
\EndFor
\end{algorithmic}
\end{algorithm}

\subsubsection{Implementation Details}
% The overall algorithm framework is illustrated in Algo. 1. We used GoogLeNet as the image feature extractor for both the SumMe and TVSum datasets, while Blip \cite{li2022blip} was selected as the feature extractor specifically for the VideoXum dataset. During the training phase, we used AdamW \cite{loshchilov2017decoupled} as the optimizer, with a dropout rate of 0.4 and an L2 regularization term set to 0.01, to effectively mitigate overfitting. In line with previous research, we used KTS to segment videos \cite{potapov2014category} and applied the Knapsack algorithm \cite{pisinger1999core} to solve the knapsack problem, using the scores generated by our model as input. This approach enabled us to derive the optimal summary solution.

We have implemented and conducted our experiments following a well-defined and reproducible framework. As illustrated in Algorithm 1, we have utilized GoogLeNet as the image feature extractor for the SumMe and TVSum datasets, while Blip \cite{li2022blip} has been specifically employed for the VideoXum dataset to better capture visual representations. To optimize the training process, we have adopted AdamW \cite{loshchilov2017decoupled} as the optimizer, setting the dropout rate to 0.4 and an L2 regularization term of 0.01 to effectively mitigate overfitting and improve generalization. In line with previous research, we have applied Kernel Temporal Segmentation (KTS) \cite{potapov2014category} to segment videos and utilized the Knapsack algorithm \cite{pisinger1999core} to solve the knapsack problem, leveraging the scores generated by our model to determine the optimal video summary solution. This methodological approach has ensured a structured and efficient summarization process.

Furthermore, we have conducted all experiments on a single NVIDIA RTX 3090 GPU, demonstrating the computational efficiency of our method. The training has been completed within 6 hours across all dataset splits, including SumMe, TVSum, and QFVS, which contain dozens of videos, as well as the large-scale VideoXum dataset with thousands of videos. These details emphasize the efficiency and scalability of our approach, making it feasible for real-world applications, including deployment on resource-constrained devices.

\subsection{Performance Comparison}

\subsubsection{Performance on Traditional Dataset}

\begin{table*}
	\centering
	\begin{threeparttable}
		\renewcommand\arraystretch{1.5}
		\caption{The Performance Comparison Between the LGRLN with Other Baselines in SumMe and TVSum Datasets.}
		\label{TAB2}
		\setlength{\tabcolsep}{4pt}{
\begin{tabular}{cccccclcccclcccc}
\hline \hline
  \multirow{2}{*}{Type} &
  \multirow{2}{*}{Model} &
  \multicolumn{4}{c}{SumMe} &
   &
  \multicolumn{4}{c}{TVSum} &
   \\ \cline{3-6} \cline{8-11} 
 &
   &
  F1 score $\uparrow$ &
  Kendall’s $\tau$ $\uparrow$&
  Spearman’s $\rho$ $\uparrow$&
   &
   &
  F1 score $\uparrow$&
  Kendall’s $\tau$ $\uparrow$&
  Spearman’s $\rho$ $\uparrow$&

   \\ \hline
\multirow{1}{*}{Human\cite{intro8}} &
 - &
  54.0 &
  0.21 &
  0.21 &
   &
   &
  54.0 &
  0.18 &
  0.20 &
   &
\\ \hline
\multirow{7}{*}{Others} &
 SumTransfer \cite{zhang2016summary} &
  40.9 &
  - &
  - &
   &
   &
  - &
  - &
  - &
   &\\
 &
 SUM-DeepLab \cite{rochan2018video} &
  48.8 &
  - &
  - &
   &
   &
  58.4 &
  - &
  - &
   & \\
 &
 dqqLSTM \cite{zhang2016video} &
  38.6 &
  0.04 &
  0.06 &
   &
   &
  54.7 &
  - &
  - &
   & \\
 &
 DR-DSN \cite{zhou2018deep} &
  42.5 &
  0.05 &
  0.05 &
   &
   &
  58.1 &
  0.02 &
  0.03 &
   & \\
 &
 HSA-RNN \cite{zhao2018hsa} &
  42.3 &
  0.06 &
  0.07 &
   &
   &
  58.7 &
  0.08 &
  0.09 &
   & \\
\hline
\multirow{3}{*}{GNN} &
  VideoSAGE \cite{chaves2024videosage} &
  46.0 &
  0.12 &
  0.16 &
   &
   &
  58.2 &
  0.30 &
  0.42 &
   & \\
 &
  RSGN \cite{intro8}  &
  45.0 &
  0.08 &
  0.09 &
   &
   &
  60.1 &
  0.08 &
  0.09 &
   &  \\
 &
  SumGraph \cite{park2020sumgraph}  &
  51.4 &
  - &
  - &
   &
   &
  \textbf{63.9} &
  0.09 &
  0.14 &
   &  \\
 &
  LGRLN(ours) &
  \textbf{54.7} &
\textbf{0.14} &
  \textbf{0.19} &
   &
   &
  58.3 &
  \textbf{0.30} &
  \textbf{0.43} &
   &  \\
  \hline \hline
\end{tabular}
}
\end{threeparttable}
\end{table*}

We conducted comprehensive performance comparisons on the SumMe, TVSum, and VideoXum datasets between our model with current mainstream baselines, as shown in Table 1. In the SumMe dataset, human performance sets a high benchmark with an F1 score of 54.0 and correlation metrics (Kendall's $\tau$ and Spearman's $\rho$) both at 0.21. Among non-GNN models, the performance varies, with F1 scores ranging from 38.6 to 55.6. dqqLSTM performs the lowest in this group, while A2Summ and PGL-Sum reach the highest F1 scores but with relatively low correlation values. GNN-based models generally show stronger results, with SumGraph and LGRLN (ours) standing out. LGRLN nearly matches human performance with an F1 score of 54.7 and achieves the highest correlation metrics among all models (Kendall's $\tau$ = 0.14, Spearman's $\rho$ = 0.19), indicating its effectiveness in summarization tasks within this dataset. Compared to VideoSAGE, our LGRLN model supports query-guided multimodal summarization by integrating textual inputs, employs a bi-threshold graph convolution for more precise semantic filtering, and enables personalized summary generation tailored to user intent.

In the TVSum dataset, non-GNN models show improved F1 scores compared to their SumMe results, with SUM-DeepLab, A2Summ, and PGL-Sum achieving the highest scores, particularly PGL-Sum, which also leads in correlation metrics among non-GNN models. GNN-based models further demonstrated the superiorities in this dataset, with LGRLN (ours) achieving one of the highest F1 scores at 58.7, and strong correlation metrics (Kendall's $\tau$ = 0.29, Spearman's $\rho$ = 0.41), closely following the highest values observed. This underscores the strong performance of GNN-based approaches, particularly LGRLN (ours), in summarization tasks for TVSum.

\begin{table}

	\centering
	\begin{threeparttable}
		\renewcommand\arraystretch{1.5}
		\caption{The Performance Comparison in VideoXum Dataset.}
		\label{TAB3}
		\setlength{\tabcolsep}{3pt}{
\begin{tabular}{cccccclcccclcccc}
\hline \hline
  &
  \multirow{2}{*}{Model} &
  \multicolumn{4}{c}{VideoXum} &
   \\ \cline{3-6} 
 &
   &
  F1 score $\uparrow$&
  Kendall’s $\tau$ $\uparrow$&
  Spearman’s $\rho$ $\uparrow$&
   &
   
   \\ \hline
    &
 Human \cite{intro9} &
  33.8 &
  0.305 &
  0.336 &
   &
   \\ \hline
 &
 Frozen-BLIP \cite{intro9} &
  16.1 &
  0.008 &
  0.011 &
   &
   \\
 &
 VSUM-BLIP \cite{intro9} &
  23.1 &
  0.185 &
  0.246 &
   &
    \\
 &
  VTSUM-BLIP \cite{intro9} &
  22.7 &
  0.176 &
  0.232 &
   &
    \\ \hline
 &
  LGRLN(ours) &
  \textbf{32.1} &
  \textbf{0.198} &
  \textbf{0.262} &
   &
  \\
  \hline \hline
\end{tabular}
}
\end{threeparttable}
\end{table}
\subsubsection{Performance on Large Scale VideoXum Dataset}

As shown in Table \ref{TAB3}, ours LGRLN stands out as the most effective model in the VideoXum dataset, achieving the highest F1 score of 32.1 and the best correlation metrics with human judgments (Kendall's $\tau$ = 0.198, Spearman's $\rho$ = 0.262). Notably, LGRLN's use of graph-based representation effectively captures both temporal and contextual relationships within video frames, allowing for a more comprehensive understanding of the video content. Additionally, the bi-threshold graph convolution mechanism incorporated in LGRLN enables the model to intelligently filter and prioritize relevant information from neighboring frames, reducing noise and enhancing the precision of the summaries. These design choices contribute to LGRLN's ability to outperform other models, such as VSUM-BLIP and VTSUM-BLIP, which show moderate performance with F1 scores of 23.1 and 22.7, respectively. In contrast, the performance of Frozen-BLIP significantly decline with an F1 score of 16.1 and very low correlation scores, indicating its limited capacity to effectively summarize video content. The results suggest that LGRLN's advanced graph representation and convolution mechanisms are crucial for its superior performance in video summarization tasks.

\subsubsection{Performance on Multimodal Dataset}

\begin{table*}
	\centering
	\begin{threeparttable}
		\renewcommand\arraystretch{1.5}
		\caption{The Performance Comparison Between the LGRLN with Other Baselines in QFVS Dataset.}
		\label{TAB4}
		\setlength{\tabcolsep}{4pt}{
\begin{tabular}{ccccclccclccclccclccclccc}
\hline \hline
  &
  \multirow{2}{*}{Model} &
  \multicolumn{4}{c}{Vid 1} &
  \multicolumn{4}{c}{Vid 2} &
  \multicolumn{4}{c}{Vid 3} &
  \multicolumn{4}{c}{Vid 4} &
  \multicolumn{3}{c}{Average} &
   \\ \cline{3-5} \cline{7-9} \cline{11-13} \cline{15-17} \cline{19-21} 
 &
   &
  P $\uparrow$ &
  R $\uparrow$ &
  F1 $\uparrow$&
  &
  P $\uparrow$&
  R $\uparrow$&
  F1 $\uparrow$&
  &
  P $\uparrow$&
  R $\uparrow$&
  F1 $\uparrow$&
  &
  P $\uparrow$&
  R $\uparrow$&
  F1 $\uparrow$&
  &
  P $\uparrow$&
  R $\uparrow$&
  F1 $\uparrow$&
   \\ \hline
    & SeqDPP \cite{gong2014diverse}
    & 53.43
    & 29.81
    & 36.59 &
    & 44.05
    & 46.65 
    & 43.67 &
    & 49.25 
    & 17.44 
    & 25.26 &
    & 11.14 
    & 63.49 
    & 18.15 &
    & 39.47 
    & 39.35 
    & 30.92 & \\
    & SH-DPP \cite{sharghi2016query}
    & 50.56 
    & 29.64 
    & 35.67 &
    & 42.13 
    & 46.81 
    & 42.72 &
    & 51.92 
    & 29.24 
    & 36.51 &
    & 11.51 
    & 62.88 
    & 18.62 &
    & 39.03 
    & 42.14
    & 33.38 & \\
    & QC-DPP \cite{sharghi2017query}
    & 49.86 
    & 53.38 
    & 48.68 &
    & 33.71 
    & 62.09 
    & 41.66 &
    & 55.16  
    & 62.40  
    & 56.47 &
    & 21.39  
    & 63.12  
    & 29.96 &
    & 40.03 
    & 60.25 
    & 44.19 & \\
    & TPAN \cite{zhang2018query}
    & 49.66 
    & 50.91  
    & 48.74 &
    & 43.02 
    & 48.73   
    & 45.30 &
    & 58.73   
    & 56.49   
    & 56.51 &
    & 36.70   
    & 35.96   
    & 33.64 &
    & 47.03  
    & 48.02  
    & 46.05 & \\
    & CHAN  \cite{xiao2020convolutional}
    & 54.73 
    & 46.57   
    & 49.14 &
    & 45.92  
    & 50.26    
    & 46.53 &
    & 59.75   
    & 64.53   
    & 58.65 &
    & 25.23   
    & 51.16    
    & 33.42 &
    & 46.40  
    & 53.13
    & 46.94 & \\ \hline
    & LGRLN(ours)
    & \textbf{60.79}
    & 43.69   
    & \textbf{49.72} &
    & \textbf{51.91 } 
    & 59.27    
    & \textbf{54.94} &
    &\textbf{ 70.09 }  
    & 51.44   
    & \textbf{59.03} &
    & 30.28   
    & 60.66    
    & \textbf{39.97} &
    & \textbf{53.27}  
    & 53.76 
    & \textbf{50.91} & 
  \\
  \hline \hline
\end{tabular}
}
\end{threeparttable}
\end{table*}

The QFVS dataset is a multimodal resource designed for natural language-guided video summarization. As presented in Table \ref{TAB4}, LGRLN outperforms other models, achieving the highest average F1 score of 50.91, with a precision of 53.27 and a recall of 53.76. This superior performance is consistent across all four video scenarios, where LGRLN maintains a strong balance between precision and recall, resulting in highly effective summarization outcomes. LGRLN's strong performance can be attributed to its advanced graph-based representation and bi-threshold graph convolution mechanism, enabling it to capture and integrate temporal and contextual dependencies within video data effectively. These features allow LGRLN to more effectively distinguish and prioritize relevant information from video frames, resulting in more accurate and contextually relevant summaries. In contrast, models like QC-DPP, though somewhat effective with an average F1 score of 44.19, struggle to maintain consistency and accuracy, especially when handling the variability of video content. Models such as SeqDPP and SH-DPP, with average F1 scores of 30.92 and 33.38, respectively, underscore the difficulty of achieving high recall without compromising precision. LGRLN's consistent performance across various metrics highlights the effectiveness of its graph-based architecture and convolutional mechanisms.

\subsubsection{Comparison with Large Language Modeling Methods}

\begin{table*}

	\centering
	\begin{threeparttable}
		\renewcommand\arraystretch{1.5}
		\caption{Comparison of performance and computational costs with the large language model on various datasets.}
		\label{llm}
		\setlength{\tabcolsep}{3pt}{
\begin{tabular}{ccccccccccccccccccccc}
\hline \hline
  \multirow{2}{*}{Model} &
  \multicolumn{3}{c}{SumMe} &&
  \multicolumn{3}{c}{TVSum} &&
  \multicolumn{3}{c}{VideoXum} &&
  \multirow{2}{*}{Parameters} &
  \multirow{2}{*}{Training environment}
   \\ \cline{2-4} \cline{6-8} \cline{10-12} 
   &
  F1 $\uparrow$&
  $\tau$ $\uparrow$&
  $\rho$ $\uparrow$&&
  F1 $\uparrow$&
  $\tau$ $\uparrow$&
  $\rho$ $\uparrow$&&
  F1 $\uparrow$&
  $\tau$ $\uparrow$&
  $\rho$ $\uparrow$&&
   &
   
   \\ \hline
  V2Xum-LLaMA-7B~\cite{vlm_videoxum} &
  - & 0.222 & 0.293 &&
  - & 0.296 & 0.378 && 
  29.0 & 0.204 & 0.298 &&
  7B & 8 x A100 &\\ 
  
  V2Xum-LLaMA-13B~\cite{vlm_videoxum} &
  - & - & - && 
  - & - & - &&
  31.6 & 0.200 & 0.276 &&
  13B & 8 x A100 &\\ 

  M3SUM(SP)~\cite{vlm_m3sum} &
  43.6 & - & - && 
  56.9 & - & - &&
  - & - & - &&
  over 20B & - &\\ 

  M3SUM(PCoT)~\cite{vlm_m3sum} &
  41.9 & - & - && 
  57.6 & - & - &&
  - & - & - &&
   over 20B & - &\\ 
  
  \hline
   LGRLN(ours) &
  54.7 & 0.14 & 0.19 && 
  58.3 & 0.30 & 0.43 &&
  32.1 & 0.198 & 0.262 &&
  3M & 1 x RTX3090 &\\ 
  \hline \hline
\end{tabular}
}
\end{threeparttable}
\end{table*}

To demonstrate the balanced performance and cost advantages of LGRLN over LLM based methods, we compared our proposed method with LLM based video summarization methods in Table \ref{llm}. LGRLN shows significant advantages in both performance and computational efficiency over large video models (LVMs) like VideoXum-LLaMA-7B and VideoXum-LLaMA-13B. Although it has only 3M parameters, LGRLN achieves comparable or even superior performance, reducing the parameter count by approximately 87.8\% to 91.7\%. This significant reduction in parameters leads to lower computational and storage demands, substantially cutting training and inference time. For example, LGRLN can be trained and run on a single RTX3090 GPU, while the larger models require 8 A100 GPUs. In terms of performance, LGRLN excels with F1 scores of 54.7 on the SumMe dataset and 58.3 on the TVSum dataset, outperforming methods like M3SUM(SP) and M3SUM(PCoT), which have F1 scores of 43.6 and 41.9, respectively. On the VideoXum dataset, LGRLN outperforms VideoXum-LLaMA-7B and VideoXum-LLaMA-13B, with F1 scores of 32.1, surpassing their scores of 29.0 and 31.6, respectively.

\subsubsection{Convergence Stability and Generalization Analysis}
\begin{figure}
	\centering
	\includegraphics[width=1.0\linewidth]{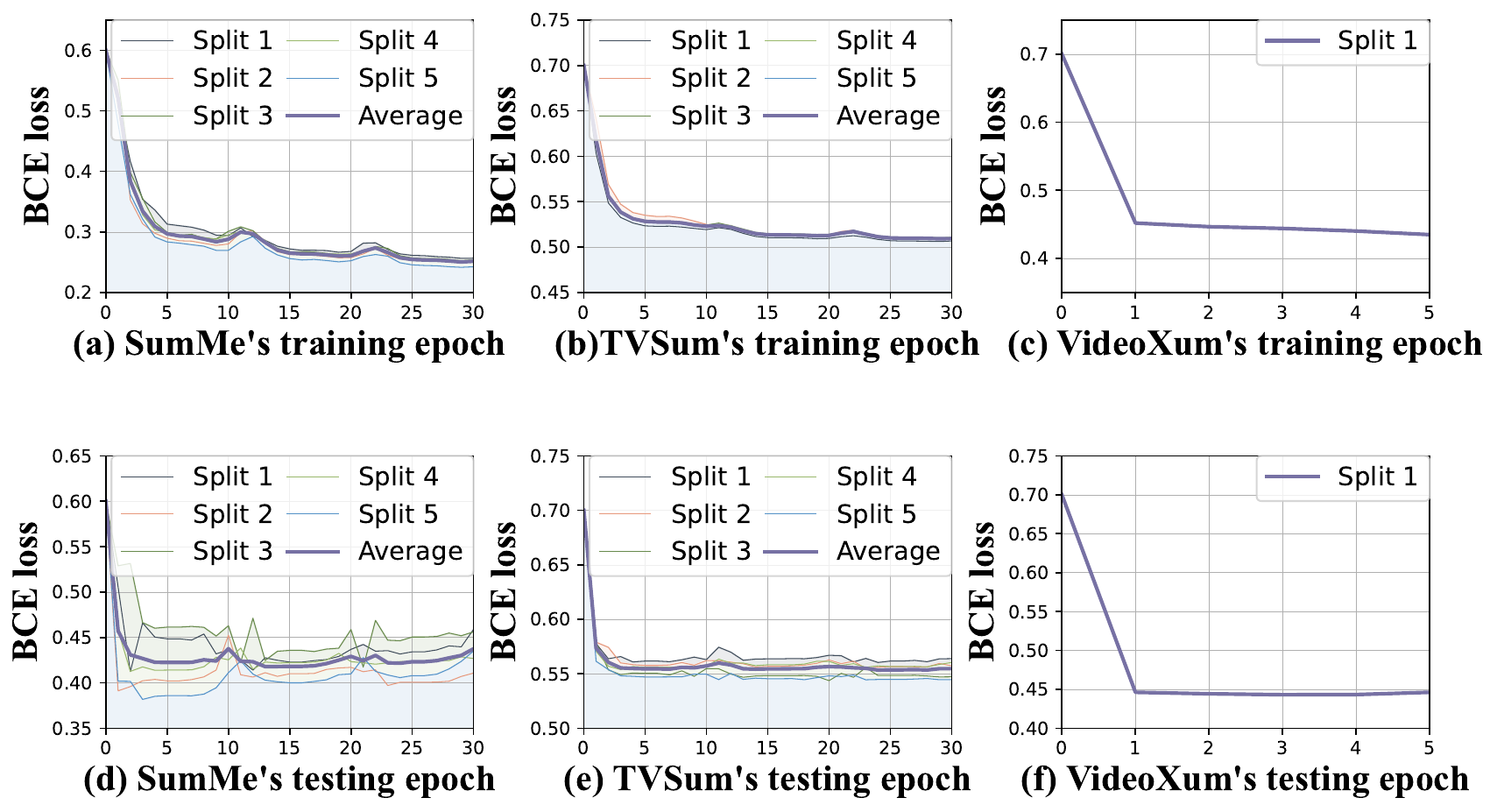}
	\caption{Learning convergence curves on mainstream datasets.}
	\label{cr}
\end{figure}

\begin{figure}
	\centering
	\includegraphics[width=1.0\linewidth]{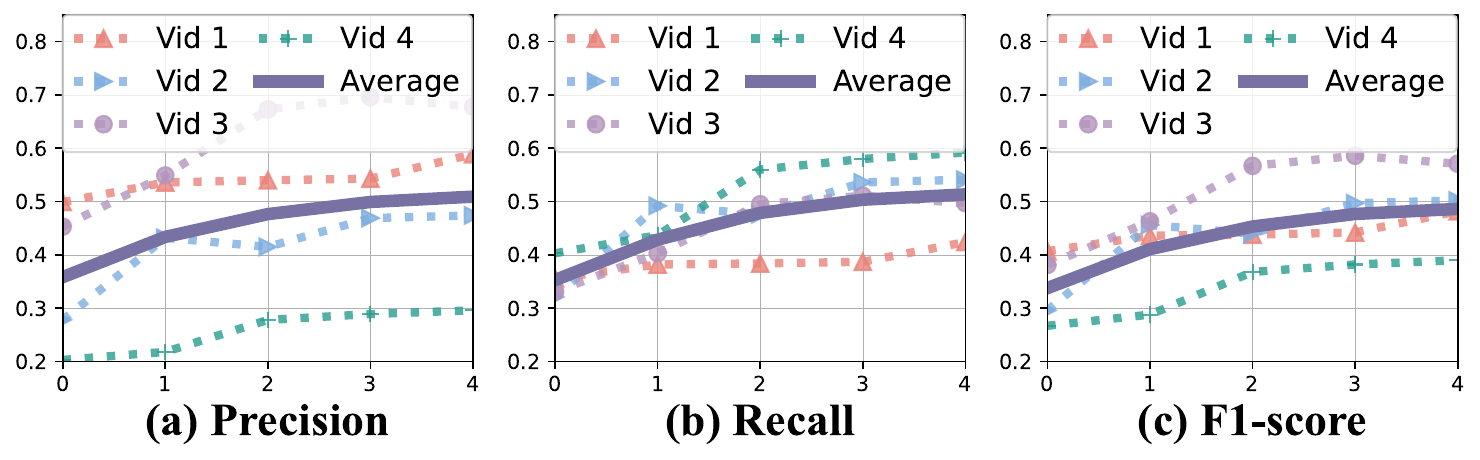}
	\caption{Learning convergence curves on QFVS dataset.}
	\label{cr}
\end{figure}

The learning curves in Fig. \ref{cr} depict the training and testing convergence of the LGRLN model on the SumMe, TVSum, and VideoXum datasets. In the SumMe and TVSum datasets, the training BCE loss decreases steadily over 30 epochs, indicating effective learning and convergence with minimal variance across different splits. This consistency across splits indicates robust model performance and stable convergence. The test BCE loss shows some fluctuations, particularly in the early epochs, possibly due to variations in the validation sets or the inherent complexity of the datasets. Despite these fluctuations, the test loss stabilizes as training progresses, confirming the model’s ability to generalize well across different splits. In contrast, the VideoXum dataset exhibits rapid convergence in the first few epochs, with both training and test losses quickly reaching low values and stabilizing, suggesting either a simpler dataset or a more effective model fit. Overall, the learning curves indicate that the LGRLN model achieves efficient and stable convergence across diverse video summarization tasks, with minor fluctuations mainly in more complex datasets like SumMe and TVSum.

\subsection{Ablation Study}

\subsubsection{The Effectiveness of Each Model Components}
Table \ref{TAB5} presents the impact and trends of different components on model performance. The baseline achieves an F1-max of 47.1 and F1-mean of 16.9 on SumMe, and 78.3 and 56.9, respectively. For single modules, both BCE and Temb bring clear improvements on SumMe, with BCE performing slightly better, while GBT shows only minor gains. On TVSum, all three modules improve F1-max, with Temb reaching the highest value of 80.8, but the F1-mean of all single modules remains below the baseline, with values of 54.6, 55.6, and 55.6. For pairwise combinations, BCE combined with GBT achieves the best results on SumMe with F1-max of 54.1 and F1-mean of 22.1, followed by BCE combined with Temb, while Temb combined with GBT shows limited improvement. On TVSum, all three combinations surpass the baseline in F1-max, with BCE and Temb achieving the highest at 81.1, but none of the combinations exceed the baseline in F1-mean, where BCE and GBT perform best at 56.8. The complete model LGRLN achieves the best performance across all metrics on both datasets, reaching 55.9 and 23.1 on SumMe, and 83.0 and 59.5 on TVSum. These results not only outperform all pairwise combinations but also make LGRLN the only configuration that clearly surpasses the baseline in F1-mean on TVSum, demonstrating that the deep integration of multiple components provides significant synergistic effects and enhanced stability.

\subsubsection{The Effectiveness of Different Hyper-parameters}
\begin{table}
	\centering
	\begin{threeparttable}
		\renewcommand\arraystretch{1.3}
		\caption{Ablation Study on Different Model Components.}
		\label{TAB5}
		\setlength{\tabcolsep}{3pt}{
\begin{tabular}{cccccccccccccccc}
\hline \hline
  \multirow{2}{*}{Model} &
  \multicolumn{3}{c}{SumMe} &&
  \multicolumn{3}{c}{TVSum} &
   \\ \cline{3-4} \cline{7-8} 
 &
   &
  F1-max $\uparrow$&
  F1-mean $\uparrow$&
   &
   &
  F1-max $\uparrow$&
  F1-mean $\uparrow$&
   
   \\ \hline
     Baseline &&
  47.1 &
  16.9 &
   &
   &
  78.3 &
  56.9 &\\ \hline
 BCE &&
  53.9 &
  20.5 &
   &
   &
  80.0 &
  54.6 &\\
  Temb &&
  52.2 &
  19.9 &
   &
   &
  80.8 &
  55.6 &\\
  GBT &&
  47.7 &
  17.2 &
   &
   &
  79.9 &
  55.6 &\\
  BCE+Temb &&
  53.9 &
  21.0 &
   &
   &
  81.1 &
  56.3 &\\
  BCE+GBT &&
  54.1 &
  22.1 &
   &
   &
  80.1 &
  56.8 &\\
  Temb+GBT &&
  52.3 &
  19.9 &
   &
   &
  80.2 &
  55.9 & \\ \hline

   LGRLN(ours) &&
  \textbf{55.9}&
  \textbf{23.1} &
   &
   &
  \textbf{83.0} &
  \textbf{59.5}&\\
  \hline \hline
\end{tabular}
}
\end{threeparttable}
\end{table}

\begin{table}

	\centering
	\begin{threeparttable}
		\renewcommand\arraystretch{1.5}
		\caption{Ablation Study on Different Weight Hyper-parameters.}
		\label{TAB6}
		\setlength{\tabcolsep}{4pt}{
\begin{tabular}{cccccclcccclcccc}
\hline \hline
  \multicolumn{2}{c}{Parameters} &
  \multicolumn{3}{c}{SumMe} &&
  \multicolumn{3}{c}{TVSum} &
   \\ \cline{1-2} \cline{4-5} \cline{8-9} 
 a & b
   & &
  F1-max $\uparrow$&
  F1-mean $\uparrow$&
   &
   &
  F1-max $\uparrow$&
  F1-mean $\uparrow$&
   
   \\ \hline
 0.000 & 0.500
 &&
  54.6 &
  16.8 &
   &
   &
  62.8 &
  42.6 &\\
 0.007 & 0.187
 &&
  52.9 &
  20.0 &
   &
   &
  71.7 &
  48.3 &\\
  0.020 & 0.160
  &&
  44.8 &
  19.4 &
   &
   &
  77.5 &
  53.9 &\\
  0.033 & 0.133
  &&
  52.3 &
  22.2 &
   &
   &
  72.3 &
  49.5 &\\ 
  0.070 & 0.100
  &&
  50.4 &
  22.8 &
   &
   &
  74.8 &
  52.0 &\\ 
  \hline \hline
\end{tabular}
}
\end{threeparttable}
\end{table}
\begin{figure*}
	\centering
	\includegraphics[scale=0.55]{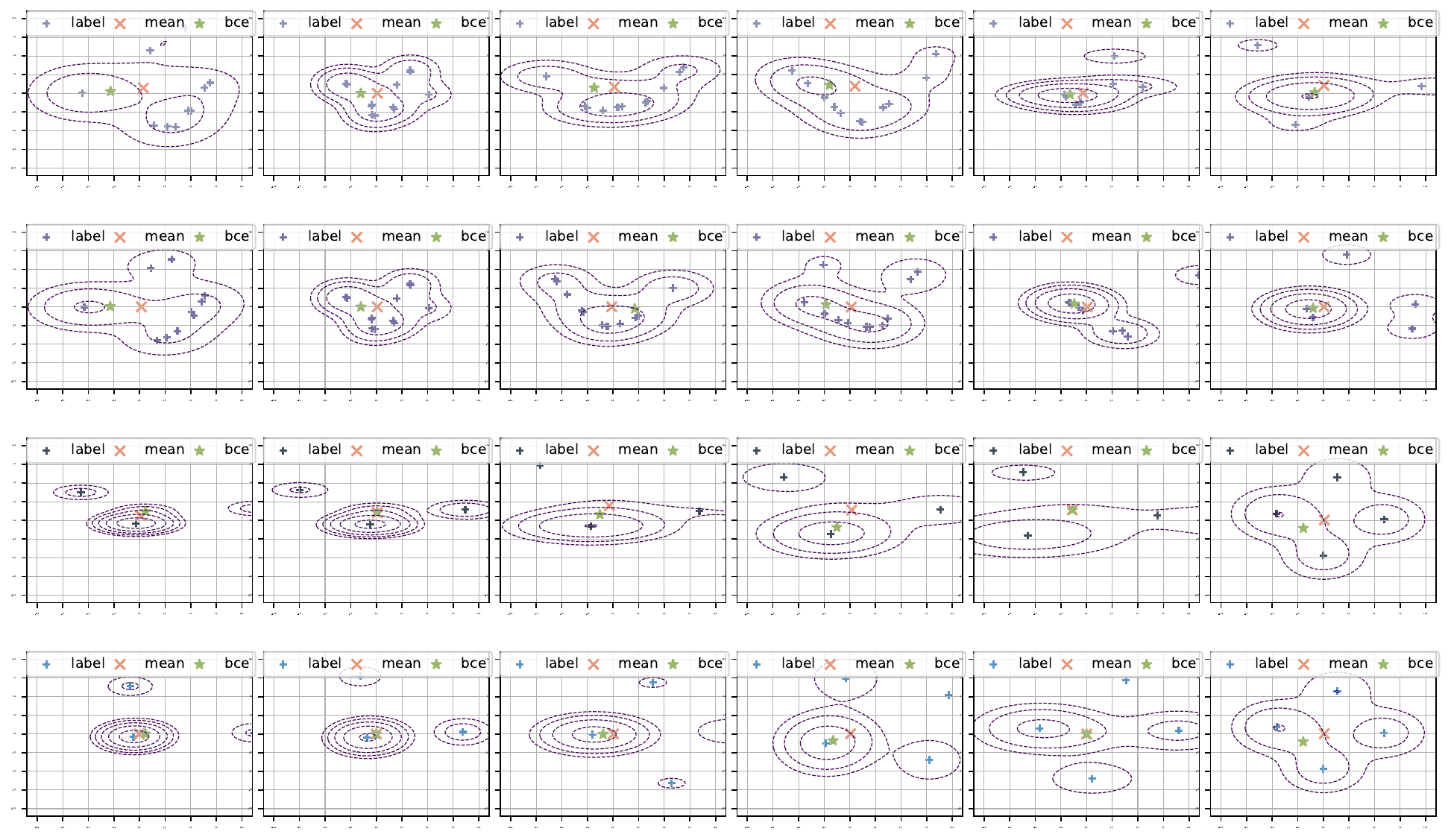}
% 	[height=8cm,width=18cm] [scale=0.78]
	\caption{The results of manually annotated summary labels for several videos in the TVSum and SumMe datasets after ISOMAP dimensionality reduction and PCA dimensionality reduction.}
	\label{gts}
\end{figure*}
\begin{figure}
	\centering
	\includegraphics[width=9cm,height=5cm]{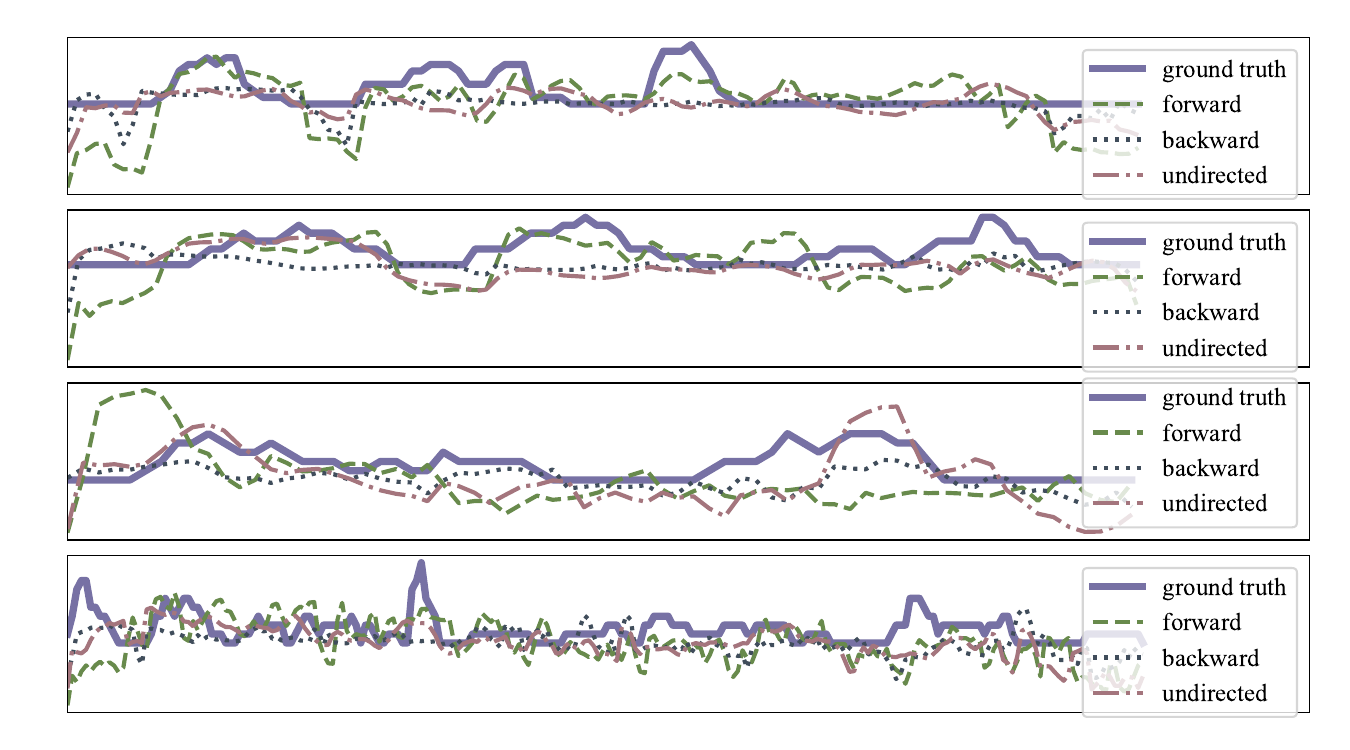}
	\caption{Summary scores from the ground truth and the outputs of graph channels for different videos. Each graph channel captures unique temporal and contextual relationships.}
	\label{channel}
\end{figure}

The ablation in Table \ref{TAB6} examines how the weight parameters $a$ and $b$ affect performance. On SumMe, reducing the difference between $b$ and $a$ generally increases F1-mean. For example, $a$=0.070 and $b$=0.100 gives F1-mean 22.8, while $a$=0.000 and $b$=0.500 gives F1-mean 16.8. The largest difference yields the highest F1-max 54.6 but with the lowest F1-mean, indicating a trade-off between peak and average performance. On TVSum, the best results occur at intermediate differences. The setting $a$=0.020 and $b$=0.160 achieves F1-max 77.5 and F1-mean 53.9, and $a$=0.070 and $b$=0.100 remains strong with F1-max 74.8 and F1-mean 52.0. A very large difference such as $a$=0.000 and $b$=0.500 degrades performance to F1-max 62.8 and F1-mean 42.6. These observations suggest that moderate or small differences between $a$ and $b$ are preferable for TVSum, whereas SumMe exhibits a trade-off where a larger difference can raise F1-max at the cost of F1-mean. Overall, tuning $a$ and $b$ controls the entropy of the weighting distribution and mid-range settings strike a better balance between fitting a single annotation strategy and generalization across annotators.

\subsection{The Effectiveness of BCE Loss}

To explore the intricacies of BCE loss mechanisms, we conducted a visual analysis comparing two aspects: the weighted average of all ground truths in BCE loss and the simple average of ground truths used in traditional methods. To visualize these high-dimensional ground truths, we employed two dimensionality reduction techniques: Principal Component Analysis (PCA) and ISOMAP. PCA reduces dimensionality by projecting data onto the hyperplane that captures the highest variance. In contrast, ISOMAP views data as residing on a manifold, using geodesic distances and Multidimensional Scaling (MDS) with equidistant properties to effectively reduce the data's dimensionality, enabling a more comprehensive visualization.

As depicted in Fig. \ref{gts}, the outcome derived from simply averaging all ground truths frequently lies outside the confines of the confidence interval established by kernel density estimation. This discrepancy highlights the limitations of traditional averaging methods in capturing the true variability and complexity of the data. In contrast, the result attained through the BCE-weighted average tends to be more intimately aligned with a particular confidence interval, indicating a closer adherence to the underlying probability distribution. This suggests that BCE loss not only preserves critical information but also aligns the model's outputs more closely with the true data distribution, making it a more effective method for handling diverse and complex annotations in video summarization tasks. These visualizations further underscore the robustness of BCE loss in managing high variability in annotations, demonstrating its superiority over traditional averaging methods.

\subsection{Analysis of Different Graph Channel Contributions}
As illustrated in Fig. \ref{channel}, the outputs of the forward, backward, and undirected graph channels exhibit notable differences, particularly in the summary score values assigned to various frames. Each graph channel processes the video sequence from a distinct perspective, capturing temporal and contextual information unique to its directionality. For instance, the forward graph channel emphasizes the progression of frames in chronological order, while the backward graph retraces the sequence to capture contextual dependencies in reverse. The undirected graph, on the other hand, focuses on bidirectional relationships, offering a holistic view of frame correlations without enforcing a strict temporal order.

These differences indicate that each channel independently identifies keyframes that are valuable for summarization, leveraging complementary insights from the video sequence flow. By integrating the results from these three methods, the model ensures a comprehensive and balanced selection of summary frames. This multi-channel approach enhances the robustness and completeness of the final video summary, as it synthesizes diverse temporal perspectives to generate an accurate and semantically rich output.

\subsection{Temporal Dynamics Captured by Time Embedding}

\begin{figure}
	\centering
	\includegraphics[width=\linewidth]{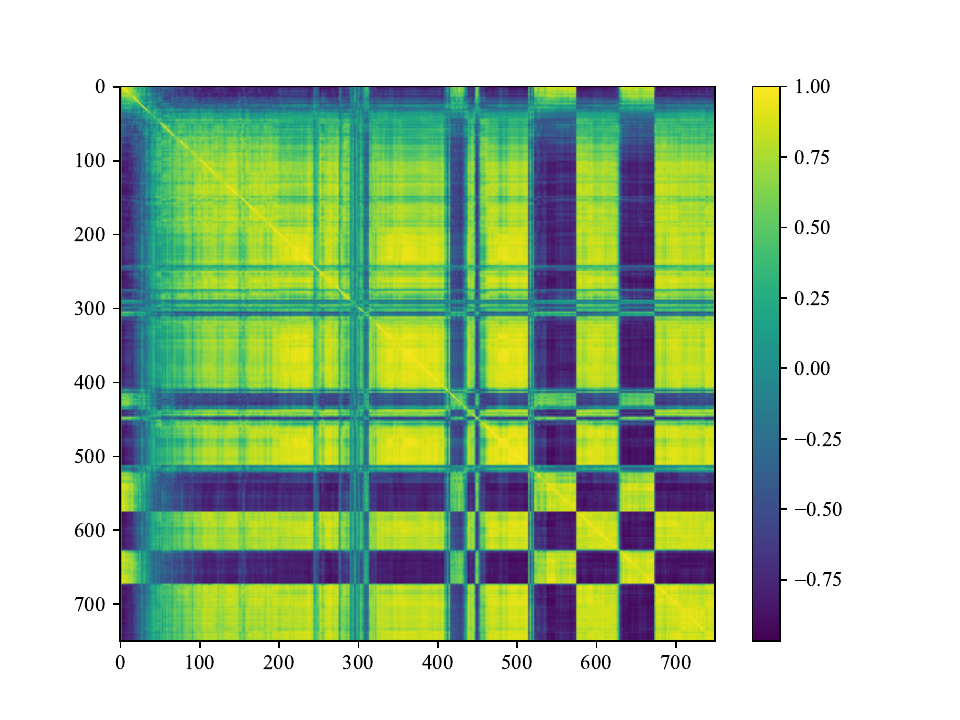}
	\caption{Correlation heatmap of time embeddings at different time points after training. Strong correlations between adjacent moments and periodic fluctuations across intervals demonstrate the time embedding's ability to capture both immediate and long-term temporal patterns.}
	\label{emb}
\end{figure}

As shown in Fig. \ref{emb}, the time embedding mechanism demonstrates a strong correlation between the embeddings of adjacent time points. This correlation exhibits periodic fluctuations across varying time intervals, indicating that the time embedding effectively captures temporal patterns and frequency characteristics within the video data. These periodic trends reflect the model's ability to encode temporal dependencies and cyclical behaviors inherent in the video sequence, such as repeated actions or transitions. The heatmap in Fig. \ref{emb} illustrates this phenomenon in detail, where the diagonal elements, representing the correlation of embeddings at consecutive time points, are notably strong. Additionally, the off-diagonal elements show periodic correlation patterns, signifying that the time embedding not only captures immediate temporal relationships but also recognizes long-term dependencies and periodic structures within the video. This ability to encode both short-term and long-term temporal features ensures that the time embedding aligns well with the sequential nature of video data.

By learning these temporal dynamics, the time embedding contributes to the model's understanding of sequential relationships and ensures that the temporal context is properly integrated into the summarization process. This capability enhances the model's ability to generate summaries that are not only temporally coherent but also sensitive to recurring patterns or significant time-based variations within the video content.

\subsection{Summary of Videos Guided by Natural Language}

\begin{figure}
	\centering
	\includegraphics[width=9cm,height=8cm]{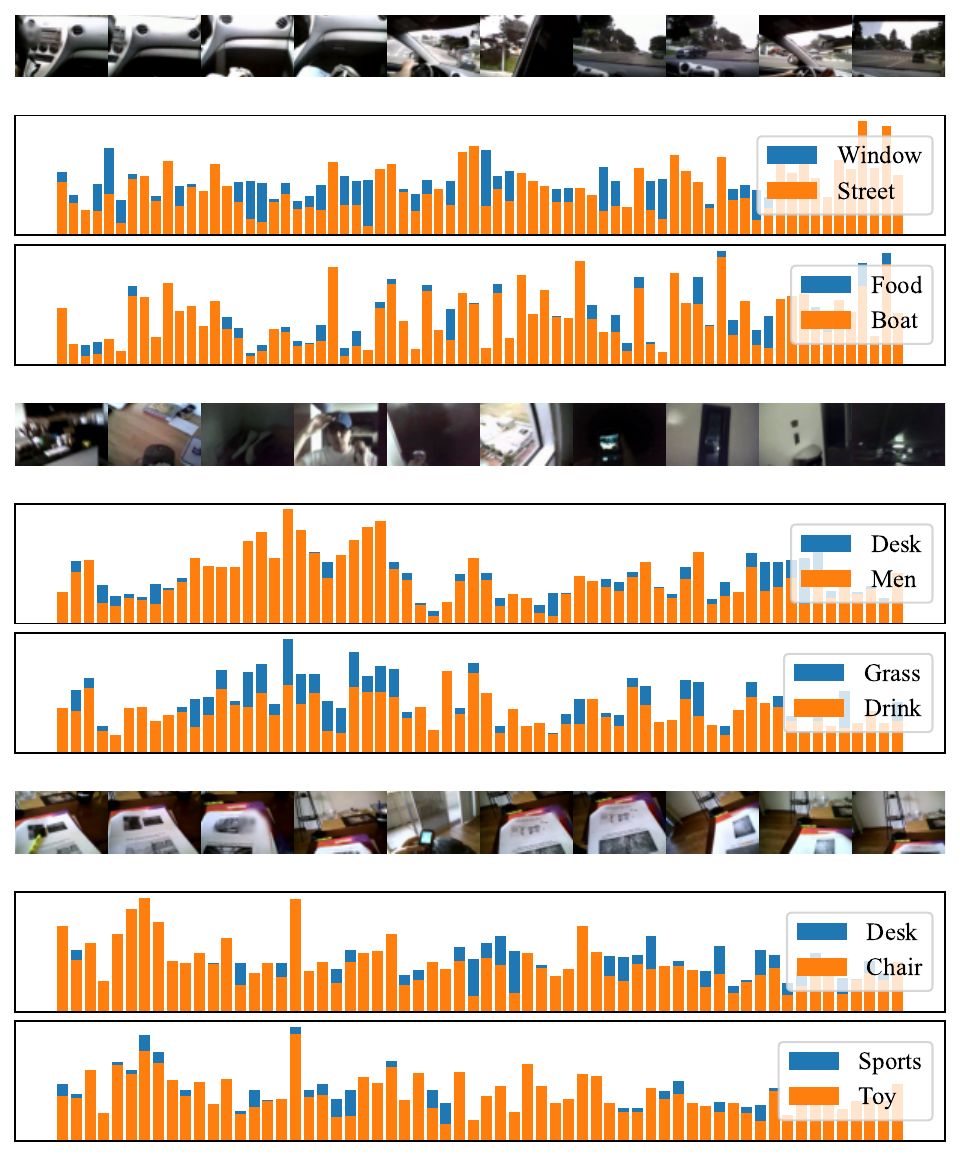}
	\caption{The video summarization results demonstrate the effectiveness of our proposed LGRLN in handling various natural language instructions. Each row presents different instructions applied to videos, with blue and orange bars indicating the frequency of key visual elements based on the prompts.}
	\label{results}
\end{figure}
\begin{table*}
	\centering
	\begin{threeparttable}
		\renewcommand\arraystretch{1.5}
		\caption{The computational cost comparison during inference in TVSum dataset.}
		\label{TAB7}
		\setlength{\tabcolsep}{12pt}{
\begin{tabular}{c|cccc|c}
\hline \hline
Model   &Kendall’s $\tau$ $\uparrow$&
  Spearman’s $\rho$ $\uparrow$& Time (ms) $\downarrow$& Parameters (MB) $\downarrow$ & Total (MB) $\downarrow$ \\ \hline
PGL-SUM \cite{apostolidis2021combining}    &0.27 &0.39 & 113.79   & 36.02   & 55.17            \\ 
A2Summ \cite{he2023align}     &0.26 &0.38 & 120.59    & 9.60    & 50.56           \\ 
VideoSAGE \cite{chaves2024videosage}  &\textbf{0.30} &0.42 & 23.55     & 3.52   & 19.27            \\ \hline
LGRLN(ours)  &\textbf{0.30} &\textbf{0.43} & \textbf{13.87}     & \textbf{2.97}   & \textbf{13.96}            \\\hline \hline
\end{tabular}}
\end{threeparttable}
\end{table*}

We present qualitative results in Fig. \ref{results} using BERT as the text encoder and visualize frame-level relevance as histograms, where the horizontal axis is the temporal order of frames and the vertical axis is the normalized relevance score; blue and orange bars correspond to two prompts. For each video, the first histogram employs prompts that actually occur in the clip, specifically Window versus Street for the first video, Desk versus Men for the second, and Desk versus Chair for the third. In these matched settings, LGRLN concentrates high scores on the correct temporal regions and suppresses unrelated frames, indicating accurate localization and effective language–vision alignment. To assess rejection behavior, the second histogram of each video uses prompts that are unrelated to the clip, specifically Food versus Boat in the first video, Grass versus Drink in the second, and Sports versus Toy in the third. Ideally the model should output uniformly low scores when no corresponding content is present, yet we observe non-negligible activations at certain time positions. These errors likely arise from spurious lexical associations between prompts, background co-occurrence patterns that partially resemble the scene, and temporal smoothing in graph message passing that propagates false positives across neighboring frames. This analysis clarifies both the strengths of the method in matched scenarios and its current limitation in rejecting off-topic prompts, and it motivates future improvements through negative-prompt training, confidence calibration with adaptive thresholds, contrastive suppression of irrelevant prompts, and prompt-conditioned gating in the relational reasoning module.

\subsection{Comparative Analysis of Computational Efficiency}

To illustrate the efficiency of our model, we performed a comparative analysis of its computational cost against several traditional video summarization models, as shown in Table \ref{TAB7}. Our LGRLN model significantly reduces computational demands, requiring only 13.96 MB of memory and 13.87 ms for inference. Compared to the PGL-SUM model, which consumes 55.17 MB of memory and takes 113.79 ms for inference, our model reduces memory usage by approximately 74.7\% and inference time by 87.8\%. Even compared to the more efficient VideoSAGE model, LGRLN reduces total memory usage by 27.5\% and inference time by 41.1\%. Additionally, our LGRLN model uses only 2.97 MB of parameters, a reduction of 91.7\% compared to PGL-SUM and 69.1\% compared to A2Summ, which require 36.02 MB and 9.60 MB of parameters, respectively. Despite this dramatic reduction in resource usage, our model maintains competitive performance among the tested models, indicating a strong correlation between the predicted summaries and the ground truth.

In summary, the LGRLN model demonstrated superiorities in producing high-quality video summaries while maintaining remarkable computational efficiency. These reductions in resource usage, up to 91.7\% in parameter count and 87.8\% in inference time, highlight its suitability for practical applications where both performance and resource optimization are crucial.

\subsection{Deployment considerations and limitations}
A practical scenario that motivates our summarization, where a user's natural-language query guides the selection of a compact set of key frames or shots to create a temporally coherent visual summary of the video. In contrast, video captioning aims to generate free-form textual descriptions for a given video and has seen rapid progress. numbers reported in this paper are measured on desktop-class hardware and are intended to indicate relative efficiency. Actual on-device deploy ability depends on factors not evaluated here, including memory footprint under the target runtime (e.g., NNAPI/Core ML), operator availability/fusion, I/O and pre/post-processing overheads, and potential quantization/distillation. While our model has only 3M parameters and shows reduced inference cost, demonstrating mobile feasibility requires hardware-specific profiling and engineering, which we leave for future work.
\section{Conclusion}\label{conclusion}
In this paper, we propose a novel Language-Guided Graph Representation Learning Network (LGRLN) to tackle the challenges of video summarization, with a focus on capturing temporal dependencies and integrating multimodal user inputs. By transforming video frames into structured graphs and integrating user-provided language instructions, our approach generates personalized and contextually relevant video summaries. Experimental results show that our method outperforms existing approaches across multiple benchmarks, particularly excelling in multimodal tasks. Furthermore, our application of a mixture Bernoulli distribution model for managing diverse annotation sets a new standard in accommodating user preferences in summary generation. Future work could involve further refinement of graph representation and the integration of additional modalities, such as audio or user interaction data, to enhance the adaptability of the summarization process. 
% use section* for acknowledgmentfv
\ifCLASSOPTIONcompsoc
  \section*{Acknowledgments}
\else
  \section*{Acknowledgment}
\fi

This work was supported in part by the National Key R\&D Program of China (2023YFA1008501) and the National Natural Science Foundation of China (NSFC) under grant 624B2049 and U22B2035.

\ifCLASSOPTIONcaptionsoff
  \newpage
\fi



\bibliographystyle{IEEEtran}
% \bibliography{tpami}

\begin{thebibliography}{10}
\providecommand{\url}[1]{#1}
\csname url@samestyle\endcsname
\providecommand{\newblock}{\relax}
\providecommand{\bibinfo}[2]{#2}
\providecommand{\BIBentrySTDinterwordspacing}{\spaceskip=0pt\relax}
\providecommand{\BIBentryALTinterwordstretchfactor}{4}
\providecommand{\BIBentryALTinterwordspacing}{\spaceskip=\fontdimen2\font plus
\BIBentryALTinterwordstretchfactor\fontdimen3\font minus \fontdimen4\font\relax}
\providecommand{\BIBforeignlanguage}[2]{{%
\expandafter\ifx\csname l@#1\endcsname\relax
\typeout{** WARNING: IEEEtran.bst: No hyphenation pattern has been}%
\typeout{** loaded for the language `#1'. Using the pattern for}%
\typeout{** the default language instead.}%
\else
\language=\csname l@#1\endcsname
\fi
#2}}
\providecommand{\BIBdecl}{\relax}
\BIBdecl

\bibitem{wenrui111}
W.~Li, W.~Han, L.-J. Deng, R.~Xiong, and X.~Fan, ``Spiking variational graph representation inference for video summarization,'' \emph{IEEE Transactions on Image Processing}, vol.~34, pp. 5697--5709, 2025.

\bibitem{Chen_02}
Z.~Chen, J.~Zhang, Z.~Lai, G.~Zhu, Z.~Liu, J.~Chen, and J.~Li, ``The devil is in the crack orientation: A new perspective for crack detection,'' in \emph{Proceedings of the IEEE/CVF International Conference on Computer Vision (ICCV)}, October 2023, pp. 6653--6663.

\bibitem{Chen_03}
Z.~Chen, Z.~Lai, J.~Chen, and J.~Li, ``Mind marginal non-crack regions: Clustering-inspired representation learning for crack segmentation,'' in \emph{Proceedings of the IEEE/CVF Conference on Computer Vision and Pattern Recognition (CVPR)}, June 2024, pp. 12\,698--12\,708.

\bibitem{zhuoyuan01}
Z.~Li, J.~Liao, C.~Tang, H.~Zhang, Y.~Li, Y.~Bian, X.~Sheng, X.~Feng, Y.~Li, C.~Gao \emph{et~al.}, ``Ustc-td: A test dataset and benchmark for image and video coding in 2020s,'' \emph{IEEE Transactions on Multimedia}, 2025.

\bibitem{zhaorui1}
R.~Zhao, R.~Xiong, J.~Zhang, Z.~Yu, S.~Zhu, L.~Ma, and T.~Huang, ``Spike camera image reconstruction using deep spiking neural networks,'' \emph{IEEE Transactions on Circuits and Systems for Video Technology}, vol.~34, no.~6, pp. 5207--5212, 2023.

\bibitem{zhaorui2}
R.~Zhao, R.~Xiong, Z.~Ding, X.~Fan, J.~Zhang, and T.~Huang, ``Mrdflow: Unsupervised optical flow estimation network with multi-scale recurrent decoder,'' \emph{IEEE Transactions on Circuits and Systems for Video Technology}, vol.~32, no.~7, pp. 4639--4652, 2021.

\bibitem{xiao01}
Z.~Xiao, Z.~Li, and W.~Jia, ``Occlusion-embedded hybrid transformer for light field super-resolution,'' in \emph{Proceedings of the AAAI Conference on Artificial Intelligence}, vol.~39, no.~8, 2025, pp. 8700--8708.

\bibitem{xiao02}
Z.~Xiao and X.~Wang, ``Event-based video super-resolution via state space models,'' in \emph{Proceedings of the Computer Vision and Pattern Recognition Conference}, 2025, pp. 12\,564--12\,574.

\bibitem{wenrui222}
W.~Li, P.~Wang, R.~Xiong, and X.~Fan, ``Spiking tucker fusion transformer for audio-visual zero-shot learning,'' \emph{IEEE Transactions on Image Processing}, vol.~33, pp. 4840--4852, 2024.

\bibitem{wenrui333}
W.~Li, Z.~Ma, L.-J. Deng, X.~Fan, and Y.~Tian, ``Neuron-based spiking transmission and reasoning network for robust image-text retrieval,'' \emph{IEEE Transactions on Circuits and Systems for Video Technology}, vol.~33, no.~7, pp. 3516--3528, 2023.

\bibitem{wenrui444}
W.~Li, P.~Wang, X.~Wang, W.~Zuo, X.~Fan, and Y.~Tian, ``Multi-timescale motion-decoupled spiking transformer for audio-visual zero-shot learning,'' \emph{IEEE Transactions on Circuits and Systems for Video Technology}, vol.~35, no.~11, pp. 10\,772--10\,786, 2025.

\bibitem{wenrui5}
W.~Li, R.~Xiong, and X.~Fan, ``Multi-layer probabilistic association reasoning network for image-text retrieval,'' \emph{IEEE Transactions on Circuits and Systems for Video Technology}, pp. 1--1, 2024.

\bibitem{wenrui6}
\BIBentryALTinterwordspacing
W.~Li, X.-L. Zhao, Z.~Ma, X.~Wang, X.~Fan, and Y.~Tian, ``Motion-decoupled spiking transformer for audio-visual zero-shot learning,'' ser. MM '23.\hskip 1em plus 0.5em minus 0.4em\relax New York, NY, USA: Association for Computing Machinery, 2023, p. 3994–4002. [Online]. Available: \url{https://doi.org/10.1145/3581783.3611759}
\BIBentrySTDinterwordspacing

\bibitem{wenrui7}
\BIBentryALTinterwordspacing
W.~Li, Z.~Ma, L.-J. Deng, P.~Wang, J.~Shi, and X.~Fan, ``Reservoir computing transformer for image-text retrieval,'' in \emph{Proceedings of the 31st ACM International Conference on Multimedia}, ser. MM '23.\hskip 1em plus 0.5em minus 0.4em\relax New York, NY, USA: Association for Computing Machinery, 2023, p. 5605–5613. [Online]. Available: \url{https://doi.org/10.1145/3581783.3611758}
\BIBentrySTDinterwordspacing

\bibitem{bai01}
H.~Bai, Z.~Zhao, J.~Zhang, Y.~Wu, L.~Deng, Y.~Cui, B.~Jiang, and S.~Xu, ``Refusion: Learning image fusion from reconstruction with learnable loss via meta-learning,'' \emph{International Journal of Computer Vision}, pp. 1--21, 2024.

\bibitem{Bai02}
H.~Bai, J.~Zhang, Z.~Zhao, Y.~Wu, L.~Deng, Y.~Cui, T.~Feng, and S.~Xu, ``Task-driven image fusion with learnable fusion loss,'' in \emph{Proceedings of the IEEE/CVF Conference on Computer Vision and Pattern Recognition (CVPR)}, June 2025, pp. 7457--7468.

\bibitem{zhang01}
X.~Zhang, J.~Ma, G.~Wang, Q.~Zhang, H.~Zhang, and L.~Zhang, ``Perceive-ir: Learning to perceive degradation better for all-in-one image restoration,'' \emph{IEEE Transactions on Image Processing}, pp. 1--1, 2025.

\bibitem{zhang02}
X.~Zhang, H.~Zhang, G.~Wang, Q.~Zhang, L.~Zhang, and B.~Du, ``Uniuir: Considering underwater image restoration as an all-in-one learner,'' \emph{IEEE Transactions on Image Processing}, vol.~34, pp. 6963--6977, 2025.

\bibitem{intro1}
A.~Mitra, S.~Biswas, and C.~Bhattacharyya, ``Bayesian modeling of temporal coherence in videos for entity discovery and summarization,'' \emph{IEEE transactions on pattern analysis and machine intelligence}, vol.~39, no.~3, pp. 430--443, 2016.

\bibitem{intro2}
M.~Sun, A.~Farhadi, B.~Taskar, and S.~Seitz, ``Summarizing unconstrained videos using salient montages,'' \emph{IEEE transactions on pattern analysis and machine intelligence}, vol.~39, no.~11, pp. 2256--2269, 2016.

\bibitem{intro3}
B.~Zhao, X.~Li, and X.~Lu, ``Hsa-rnn: Hierarchical structure-adaptive rnn for video summarization,'' in \emph{Proceedings of the IEEE conference on computer vision and pattern recognition}, 2018, pp. 7405--7414.

\bibitem{intro4}
Z.~Wei, B.~Wang, M.~Hoai, J.~Zhang, X.~Shen, Z.~Lin, R.~M{\v{e}}ch, and D.~Samaras, ``Sequence-to-segments networks for detecting segments in videos,'' \emph{IEEE transactions on pattern analysis and machine intelligence}, vol.~43, no.~3, pp. 1009--1021, 2019.

\bibitem{intro5}
M.~Jia, Y.~Wei, X.~Song, T.~Sun, M.~Zhang, and L.~Nie, ``Query-oriented micro-video summarization,'' \emph{IEEE Transactions on Pattern Analysis and Machine Intelligence}, vol.~46, no.~6, pp. 4174--4187, 2024.

\bibitem{intro6}
M.~Narasimhan, A.~Rohrbach, and T.~Darrell, ``Clip-it! language-guided video summarization,'' \emph{Advances in neural information processing systems}, vol.~34, pp. 13\,988--14\,000, 2021.

\bibitem{intro7}
W.~Zhu, J.~Lu, J.~Li, and J.~Zhou, ``Dsnet: A flexible detect-to-summarize network for video summarization,'' \emph{IEEE Transactions on Image Processing}, vol.~30, pp. 948--962, 2020.

\bibitem{intro8}
B.~Zhao, H.~Li, X.~Lu, and X.~Li, ``Reconstructive sequence-graph network for video summarization,'' \emph{IEEE Transactions on Pattern Analysis and Machine Intelligence}, vol.~44, no.~5, pp. 2793--2801, 2021.

\bibitem{intro9}
J.~Lin, H.~Hua, M.~Chen, Y.~Li, J.~Hsiao, C.~Ho, and J.~Luo, ``Videoxum: Cross-modal visual and textural summarization of videos,'' \emph{IEEE Transactions on Multimedia}, vol.~26, pp. 5548--5560, 2024.

\bibitem{intro10}
H.~Li, Q.~Ke, M.~Gong, and R.~Zhang, ``Video joint modelling based on hierarchical transformer for co-summarization,'' \emph{IEEE Transactions on Pattern Analysis and Machine Intelligence}, vol.~45, no.~3, pp. 3904--3917, 2022.

\bibitem{intro11}
X.~Li, B.~Zhao, and X.~Lu, ``A general framework for edited video and raw video summarization,'' \emph{IEEE Transactions on Image Processing}, vol.~26, no.~8, pp. 3652--3664, 2017.

\bibitem{intro12}
T.~Hussain, K.~Muhammad, W.~Ding, J.~Lloret, S.~W. Baik, and V.~H.~C. {de Albuquerque}, ``A comprehensive survey of multi-view video summarization,'' \emph{Pattern Recognition}, vol. 109, p. 107567, 2021.

\bibitem{lirevised1}
G.~Li, H.~Ye, Y.~Qi, S.~Wang, L.~Qing, Q.~Huang, and M.-H. Yang, ``Learning hierarchical modular networks for video captioning,'' \emph{IEEE Transactions on Pattern Analysis and Machine Intelligence}, vol.~46, no.~2, pp. 1049--1064, 2024.

\bibitem{lirevised2}
Y.~Ma, Z.~Zhu, Y.~Qi, A.~Beheshti, Y.~Li, L.~Qing, and G.~Li, ``Style-aware two-stage learning framework for video captioning,'' \emph{Knowledge-Based Systems}, vol. 301, p. 112258, 2024.

\bibitem{lirevised3}
M.~Tian, G.~Li, Y.~Qi, S.~Wang, Q.~Z. Sheng, and Q.~Huang, ``Rethink video retrieval representation for video captioning,'' \emph{Pattern Recognition}, vol. 156, p. 110744, 2024.

\bibitem{lirevised4}
Y.~Ma, L.~Qing, G.~Li, Y.~Qi, A.~Beheshti, Q.~Z. Sheng, and Q.~Huang, ``Retta: Retrieval-enhanced test-time adaptation for zero-shot video captioning,'' \emph{Pattern Recognition}, vol. 171, p. 112170, 2026.

\bibitem{qigraph1}
Y.~Hong, C.~Rodriguez-Opazo, Y.~Qi, Q.~Wu, and S.~Gould, ``Language and visual entity relationship graph for agent navigation,'' in \emph{Proceedings of the 34th International Conference on Neural Information Processing Systems}, ser. NIPS '20.\hskip 1em plus 0.5em minus 0.4em\relax Red Hook, NY, USA: Curran Associates Inc., 2020.

\bibitem{qigraph2}
N.~Shabani, A.~Beheshti, Y.~Qi, V.~Haghighi, S.~Moradizeyveh, and Q.~Z. Sheng, ``Trffc: Efficient traffic forecasting through adaptive spatio-temporal graph reduction,'' in \emph{Companion Proceedings of the ACM on Web Conference}, 2025, p. 2899–2902.

\bibitem{super_1}
M.~Gygli, H.~Grabner, H.~Riemenschneider, and L.~Van~Gool, ``Creating summaries from user videos,'' in \emph{Computer Vision -- ECCV 2014}, D.~Fleet, T.~Pajdla, B.~Schiele, and T.~Tuytelaars, Eds., 2014, pp. 505--520.

\bibitem{super_2}
B.~Gong, W.-L. Chao, K.~Grauman, and F.~Sha, ``Diverse sequential subset selection for supervised video summarization,'' in \emph{Proceedings of the 27th International Conference on Neural Information Processing Systems - Volume 2}, ser. NIPS'14.\hskip 1em plus 0.5em minus 0.4em\relax MIT Press, 2014, p. 2069–2077.

\bibitem{super_3}
B.~Zhao, X.~Li, and X.~Lu, ``Tth-rnn: Tensor-train hierarchical recurrent neural network for video summarization,'' \emph{IEEE Transactions on Industrial Electronics}, vol.~68, no.~4, pp. 3629--3637, 2021.

\bibitem{super_4}
K.~Zhang, K.~Grauman, and F.~Sha, ``Retrospective encoders for video summarization,'' in \emph{Computer Vision -- ECCV 2018}, V.~Ferrari, M.~Hebert, C.~Sminchisescu, and Y.~Weiss, Eds., 2018, pp. 391--408.

\bibitem{super_5}
Y.~Chen, M.~Rohrbach, Z.~Yan, S.~Yan, J.~Feng, and Y.~Kalantidis, ``Graph-based global reasoning networks,'' \emph{2019 IEEE/CVF Conference on Computer Vision and Pattern Recognition (CVPR)}, pp. 433--442, 2018.

\bibitem{super_6}
R.~Zeng, W.~Huang, M.~Tan, Y.~Rong, P.~Zhao, J.~Huang, and C.~Gan, ``Graph convolutional module for temporal action localization in videos,'' \emph{IEEE Transactions on Pattern Analysis and Machine Intelligence}, vol.~44, no.~10, pp. 6209--6223, 2022.

\bibitem{chaves2024videosage}
J.~M.~R. Chaves and S.~Tripathi, ``Videosage: Video summarization with graph representation learning,'' in \emph{Proceedings of the IEEE/CVF Conference on Computer Vision and Pattern Recognition}, 2024, pp. 2527--2534.

\bibitem{super_7}
M.~Rochan, L.~Ye, and Y.~Wang, ``Video summarization using fully convolutional sequence networks,'' in \emph{Proceedings of the European Conference on Computer Vision (ECCV)}, 2018, pp. 347--363.

\bibitem{unsupevised1}
Y.~J. Lee, J.~Ghosh, and K.~Grauman, ``Discovering important people and objects for egocentric video summarization,'' in \emph{IEEE Conference on Computer Vision and Pattern Recognition}, 2012, pp. 1346--1353.

\bibitem{unsupevised2}
P.~Mundur, Y.~Rao, and Y.~Yesha, ``Keyframe-based video summarization using delaunay clustering,'' \emph{International Journal on Digital Libraries}, vol.~6, pp. 219--232, 2006.

\bibitem{unsupevised3}
E.~Elhamifar, G.~Sapiro, and R.~Vidal, ``See all by looking at a few: Sparse modeling for finding representative objects,'' in \emph{2012 IEEE conference on computer vision and pattern recognition}.\hskip 1em plus 0.5em minus 0.4em\relax IEEE, 2012, pp. 1600--1607.

\bibitem{park2020sumgraph}
J.~Park, J.~Lee, I.-J. Kim, and K.~Sohn, ``Sumgraph: Video summarization via recursive graph modeling,'' in \emph{Computer Vision--ECCV 2020: 16th European Conference, Glasgow, UK, August 23--28, 2020, Proceedings, Part XXV 16}.\hskip 1em plus 0.5em minus 0.4em\relax Springer, 2020, pp. 647--663.

\bibitem{DSAVS}
S.-H. Zhong, J.~Lin, J.~Lu, A.~Fares, and T.~Ren, ``Deep semantic and attentive network for unsupervised video summarization,'' \emph{ACM Trans. Multimedia Comput. Commun. Appl.}, vol.~18, no.~2, 2022.

\bibitem{query1}
A.~Sharghi, B.~Gong, and M.~Shah, ``Query-focused extractive video summarization,'' in \emph{Computer Vision -- ECCV 2016}, 2016, pp. 3--19.

\bibitem{query2}
A.~Sharghi, J.~S. Laurel, and B.~Gong, ``Query-focused video summarization: Dataset, evaluation, and a memory network based approach,'' in \emph{2017 IEEE Conference on Computer Vision and Pattern Recognition (CVPR)}, 2017, pp. 2127--2136.

\bibitem{query3}
A.~Kanehira, L.~Van~Gool, Y.~Ushiku, and T.~Harada, ``Viewpoint-aware video summarization,'' in \emph{2018 IEEE/CVF Conference on Computer Vision and Pattern Recognition}, 2018, pp. 7435--7444.

\bibitem{LLM1}
D.~Kondratyuk, L.~Yu, X.~Gu, J.~Lezama, J.~Huang, G.~Schindler, R.~Hornung, V.~Birodkar, J.~Yan, M.-C. Chiu \emph{et~al.}, ``Videopoet: A large language model for zero-shot video generation,'' \emph{arXiv preprint arXiv:2312.14125}, 2023.

\bibitem{LLM2}
B.~Lin, Y.~Ye, B.~Zhu, J.~Cui, M.~Ning, P.~Jin, and L.~Yuan, ``Video-llava: Learning united visual representation by alignment before projection,'' \emph{arXiv preprint arXiv:2311.10122}, 2023.

\bibitem{LLM3}
G.~Chen, Y.-D. Zheng, J.~Wang, J.~Xu, Y.~Huang, J.~Pan, Y.~Wang, Y.~Wang, Y.~Qiao, T.~Lu \emph{et~al.}, ``Videollm: Modeling video sequence with large language models,'' \emph{arXiv preprint arXiv:2305.13292}, 2023.

\bibitem{LLM4}
K.~Li, Y.~He, Y.~Wang, Y.~Li, W.~Wang, P.~Luo, Y.~Wang, L.~Wang, and Y.~Qiao, ``Videochat: Chat-centric video understanding,'' \emph{arXiv preprint arXiv:2305.06355}, 2023.

\bibitem{LLM5}
H.~Fei, S.~Wu, M.~Zhang, M.~Zhang, T.-S. Chua, and S.~Yan, ``Enhancing video-language representations with structural spatio-temporal alignment,'' \emph{IEEE Transactions on Pattern Analysis and Machine Intelligence}, vol.~46, no.~12, pp. 7701--7719, 2024.

\bibitem{VLLM1}
H.~Fei, S.~Wu, W.~Ji, H.~Zhang, M.~Zhang, M.-L. Lee, and W.~Hsu, ``Video-of-thought: Step-by-step video reasoning from perception to cognition,'' in \emph{Proceedings of the 41st International Conference on Machine Learning}, vol. 235.\hskip 1em plus 0.5em minus 0.4em\relax PMLR, 21--27 Jul 2024, pp. 13\,109--13\,125.

\bibitem{VLLM2}
M.~Maaz, H.~Rasheed, S.~Khan, and F.~S. Khan, ``Video-chatgpt: Towards detailed video understanding via large vision and language models,'' \emph{arXiv preprint arXiv:2306.05424}, 2023.

\bibitem{VLLM3}
L.~Qian, J.~Li, Y.~Wu, Y.~Ye, H.~Fei, T.-S. Chua, Y.~Zhuang, and S.~Tang, ``Momentor: Advancing video large language model with fine-grained temporal reasoning,'' \emph{arXiv preprint arXiv:2402.11435}, 2024.

\bibitem{VLLM4}
Z.~Wang, L.~Wang, Z.~Zhao, M.~Wu, C.~Lyu, H.~Li, D.~Cai, L.~Zhou, S.~Shi, and Z.~Tu, ``Gpt4video: A unified multimodal large language model for lnstruction-followed understanding and safety-aware generation,'' in \emph{Proceedings of the 32nd ACM International Conference on Multimedia}, 2024, pp. 3907--3916.

\bibitem{vlm_videoxum}
H.~Hua, Y.~Tang, C.~Xu, and J.~Luo, ``V2xum-llm: Cross-modal video summarization with temporal prompt instruction tuning,'' \emph{arXiv preprint arXiv:2404.12353}, 2024.

\bibitem{vlm_m3sum}
H.~Wang, B.~Zhou, Z.~Zhang, Y.~Du, D.~Ho, and K.-F. Wong, ``M3sum: A novel unsupervised language-guided video summarization,'' in \emph{ICASSP 2024-2024 IEEE International Conference on Acoustics, Speech and Signal Processing (ICASSP)}.\hskip 1em plus 0.5em minus 0.4em\relax IEEE, 2024, pp. 4140--4144.

\bibitem{szegedy2015going}
C.~Szegedy, W.~Liu, Y.~Jia, P.~Sermanet, S.~Reed, D.~Anguelov, D.~Erhan, V.~Vanhoucke, and A.~Rabinovich, ``Going deeper with convolutions,'' in \emph{Proceedings of the IEEE conference on computer vision and pattern recognition}, 2015, pp. 1--9.

\bibitem{zhang2019graph}
S.~Zhang, H.~Tong, J.~Xu, and R.~Maciejewski, ``Graph convolutional networks: a comprehensive review,'' \emph{Computational Social Networks}, vol.~6, no.~1, pp. 1--23, 2019.

\bibitem{kipf2016semi}
T.~N. Kipf and M.~Welling, ``Semi-supervised classification with graph convolutional networks,'' \emph{arXiv preprint arXiv:1609.02907}, 2016.

\bibitem{velivckovic2017graph}
P.~Veli{\v{c}}kovi{\'c}, G.~Cucurull, A.~Casanova, A.~Romero, P.~Lio, and Y.~Bengio, ``Graph attention networks,'' \emph{arXiv preprint arXiv:1710.10903}, 2017.

\bibitem{hamilton2017inductive}
W.~Hamilton, Z.~Ying, and J.~Leskovec, ``Inductive representation learning on large graphs,'' \emph{Advances in neural information processing systems}, vol.~30, 2017.

\bibitem{graph_randomwalk}
H.~P. Sajjad, A.~Docherty, and Y.~Tyshetskiy, ``Efficient representation learning using random walks for dynamic graphs,'' \emph{arXiv preprint arXiv:1901.01346}, 2019.

\bibitem{hendrycks2016gaussian}
D.~Hendrycks and K.~Gimpel, ``Gaussian error linear units (gelus),'' \emph{arXiv preprint arXiv:1606.08415}, 2016.

\bibitem{cai2021graphnorm}
T.~Cai, S.~Luo, K.~Xu, D.~He, T.-y. Liu, and L.~Wang, ``Graphnorm: A principled approach to accelerating graph neural network training,'' in \emph{International Conference on Machine Learning}.\hskip 1em plus 0.5em minus 0.4em\relax PMLR, 2021, pp. 1204--1215.

\bibitem{devlin2018bert}
J.~Devlin, ``Bert: Pre-training of deep bidirectional transformers for language understanding,'' \emph{arXiv preprint arXiv:1810.04805}, 2018.

\bibitem{mclachlan2007algorithm}
G.~J. McLachlan and T.~Krishnan, \emph{The EM algorithm and extensions}.\hskip 1em plus 0.5em minus 0.4em\relax John Wiley \& Sons, 2007.

\bibitem{gygli2014creating}
M.~Gygli, H.~Grabner, H.~Riemenschneider, and L.~V. Gool, ``Creating summaries from user videos,'' in \emph{European conference on computer vision}.\hskip 1em plus 0.5em minus 0.4em\relax Springer, 2014, pp. 505--520.

\bibitem{song2015tvsum}
Y.~Song, J.~Vallmitjana, A.~Stent, and A.~Jaimes, ``Tvsum: Summarizing web videos using titles,'' in \emph{Proceedings of the IEEE conference on computer vision and pattern recognition}, 2015, pp. 5179--5187.

\bibitem{akhare2022query}
R.~Akhare and S.~Shinde, ``Query focused video summarization: A review,'' in \emph{International Symposium on Artificial Intelligence}.\hskip 1em plus 0.5em minus 0.4em\relax Springer, 2022, pp. 202--212.

\bibitem{li2022blip}
J.~Li, D.~Li, C.~Xiong, and S.~Hoi, ``Blip: Bootstrapping language-image pre-training for unified vision-language understanding and generation,'' in \emph{International conference on machine learning}.\hskip 1em plus 0.5em minus 0.4em\relax PMLR, 2022, pp. 12\,888--12\,900.

\bibitem{loshchilov2017decoupled}
I.~Loshchilov and F.~Hutter, ``Decoupled weight decay regularization,'' \emph{arXiv preprint arXiv:1711.05101}, 2017.

\bibitem{potapov2014category}
D.~Potapov, M.~Douze, Z.~Harchaoui, and C.~Schmid, ``Category-specific video summarization,'' in \emph{Computer Vision--ECCV 2014: 13th European Conference, Zurich, Switzerland, September 6-12, 2014, Proceedings, Part VI 13}.\hskip 1em plus 0.5em minus 0.4em\relax Springer, 2014, pp. 540--555.

\bibitem{pisinger1999core}
D.~Pisinger, ``Core problems in knapsack algorithms,'' \emph{Operations Research}, vol.~47, no.~4, pp. 570--575, 1999.

\bibitem{zhang2016summary}
K.~Zhang, W.-L. Chao, F.~Sha, and K.~Grauman, ``Summary transfer: Exemplar-based subset selection for video summarization,'' in \emph{Proceedings of the IEEE conference on computer vision and pattern recognition}, 2016, pp. 1059--1067.

\bibitem{rochan2018video}
M.~Rochan, L.~Ye, and Y.~Wang, ``Video summarization using fully convolutional sequence networks,'' in \emph{Proceedings of the European conference on computer vision (ECCV)}, 2018, pp. 347--363.

\bibitem{zhang2016video}
K.~Zhang, W.-L. Chao, F.~Sha, and K.~Grauman, ``Video summarization with long short-term memory,'' in \emph{Computer Vision--ECCV 2016: 14th European Conference, Amsterdam, The Netherlands, October 11--14, 2016, Proceedings, Part VII 14}.\hskip 1em plus 0.5em minus 0.4em\relax Springer, 2016, pp. 766--782.

\bibitem{zhou2018deep}
K.~Zhou, Y.~Qiao, and T.~Xiang, ``Deep reinforcement learning for unsupervised video summarization with diversity-representativeness reward,'' in \emph{Proceedings of the AAAI conference on artificial intelligence}, vol.~32, no.~1, 2018.

\bibitem{zhao2018hsa}
B.~Zhao, X.~Li, and X.~Lu, ``Hsa-rnn: Hierarchical structure-adaptive rnn for video summarization,'' in \emph{Proceedings of the IEEE conference on computer vision and pattern recognition}, 2018, pp. 7405--7414.

\bibitem{gong2014diverse}
B.~Gong, W.-L. Chao, K.~Grauman, and F.~Sha, ``Diverse sequential subset selection for supervised video summarization,'' \emph{Advances in neural information processing systems}, vol.~27, 2014.

\bibitem{sharghi2016query}
A.~Sharghi, B.~Gong, and M.~Shah, ``Query-focused extractive video summarization,'' in \emph{Computer Vision--ECCV 2016: 14th European Conference, Amsterdam, The Netherlands, October 11-14, 2016, Proceedings, Part VIII 14}.\hskip 1em plus 0.5em minus 0.4em\relax Springer, 2016, pp. 3--19.

\bibitem{sharghi2017query}
A.~Sharghi, J.~S. Laurel, and B.~Gong, ``Query-focused video summarization: Dataset, evaluation, and a memory network based approach,'' in \emph{Proceedings of the IEEE conference on computer vision and pattern recognition}, 2017, pp. 4788--4797.

\bibitem{zhang2018query}
Y.~Zhang, M.~Kampffmeyer, X.~Liang, M.~Tan, and E.~P. Xing, ``Query-conditioned three-player adversarial network for video summarization,'' \emph{arXiv preprint arXiv:1807.06677}, 2018.

\bibitem{xiao2020convolutional}
S.~Xiao, Z.~Zhao, Z.~Zhang, X.~Yan, and M.~Yang, ``Convolutional hierarchical attention network for query-focused video summarization,'' in \emph{Proceedings of the AAAI conference on artificial intelligence}, vol.~34, no.~07, 2020, pp. 12\,426--12\,433.

\bibitem{apostolidis2021combining}
E.~Apostolidis, G.~Balaouras, V.~Mezaris, and I.~Patras, ``Combining global and local attention with positional encoding for video summarization,'' in \emph{2021 IEEE international symposium on multimedia (ISM)}.\hskip 1em plus 0.5em minus 0.4em\relax IEEE, 2021, pp. 226--234.

\bibitem{he2023align}
B.~He, J.~Wang, J.~Qiu, T.~Bui, A.~Shrivastava, and Z.~Wang, ``Align and attend: Multimodal summarization with dual contrastive losses,'' in \emph{Proceedings of the IEEE/CVF conference on computer vision and pattern recognition}, 2023, pp. 14\,867--14\,878.

\end{thebibliography}

\begin{IEEEbiography}[{\includegraphics[width=1in,height=1.25in,clip,keepaspectratio]{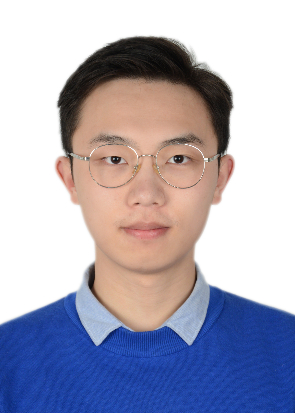}}]{Wenrui Li} received the B.S. degree from the School of Information and Software Engineering, University of Electronic Science and Technology of China (UESTC), Chengdu, China, in 2021. He is currently pursuing a Ph.D. degree in the School of Computer Science at Harbin Institute of Technology (HIT), Harbin, China. His research interests include multimedia search, joint source-channel coding, and spiking neural networks. He was supported by the National Natural Science Foundation of China (NSFC) Youth Student Basic Research Program (Doctoral Student) in 2024. He has authored or co-authored more than 30 technical articles in refereed international journals and conferences.
\end{IEEEbiography}

\begin{IEEEbiography}[{\includegraphics[width=1in,height=1.25in,clip,keepaspectratio]{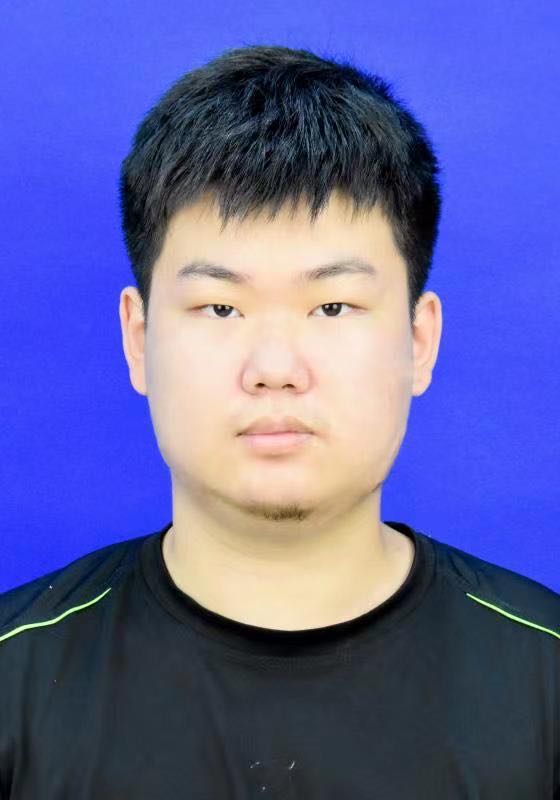}}]{Wei Han} is currently working toward the B.S. degree from the School of Computer Science, Harbin Institute of Technology (HIT), Harbin, China. His research interests include video summarization, text-3D retrieval, reinforcement learning, and LLM agent.
\end{IEEEbiography}

\begin{IEEEbiography}[{\includegraphics[width=1in,height=1.25in,clip,keepaspectratio]{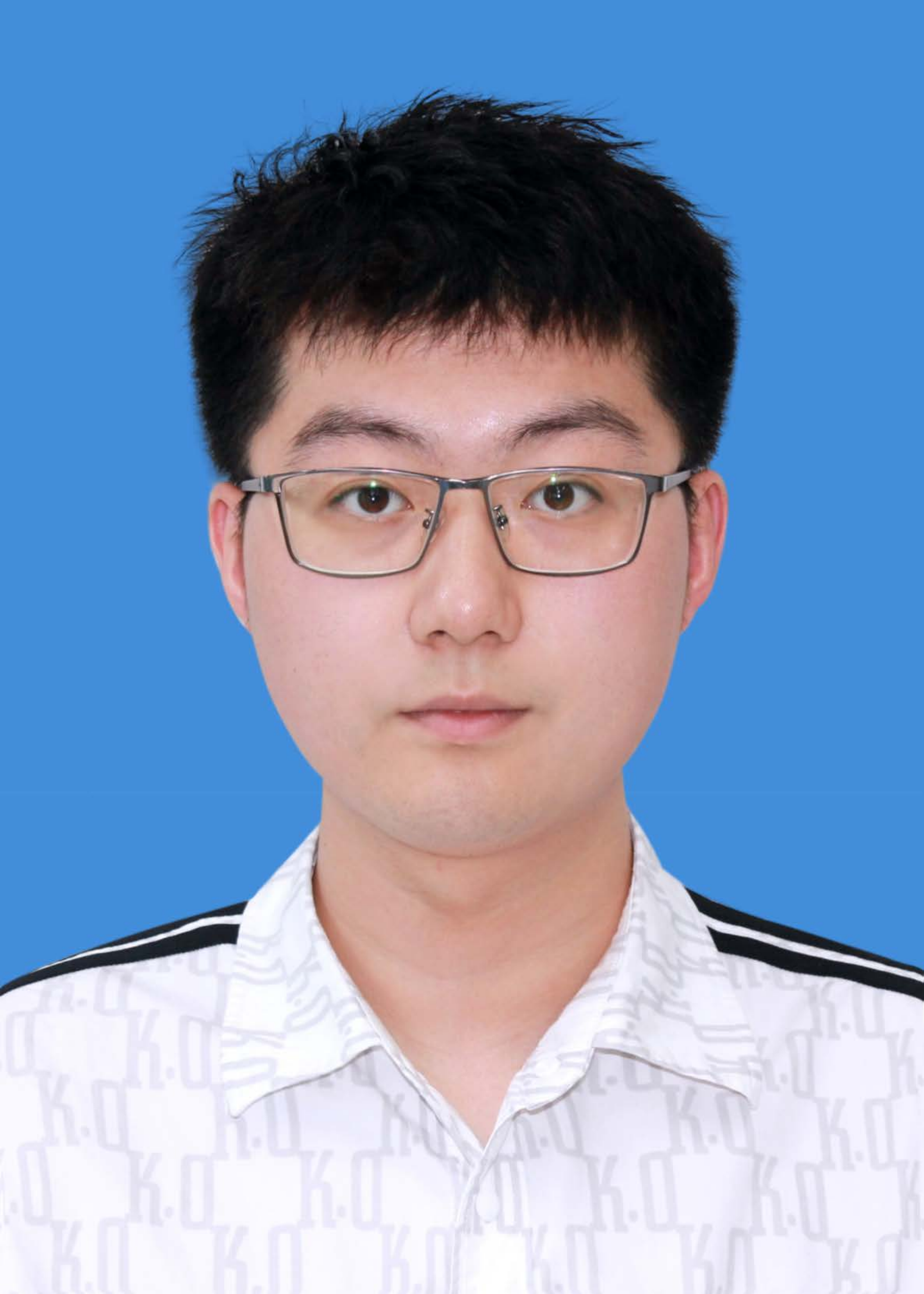}}]{Hengyu Man} received the B.S. degree in communication engineering from Beijing Jiaotong University, Beijing, China, in 2018, the M.S. degree in information and network engineering from KTH Royal Institute of Technology, Stockholm, Sweden, in 2019, and the Ph.D. degree in computer science from Harbin Institute of Technology (HIT), Harbin, China, in 2024. From 2021 to 2023, he was with Peng Cheng Laboratory, Shenzhen, China. He is currently an Associate Researcher with the Faculty of Computing, HIT. His research interests include data compression, image/video coding, and generative computer vision. 
\end{IEEEbiography}

\begin{IEEEbiography}[{\includegraphics[width=1in,height=1.25in,clip,keepaspectratio]{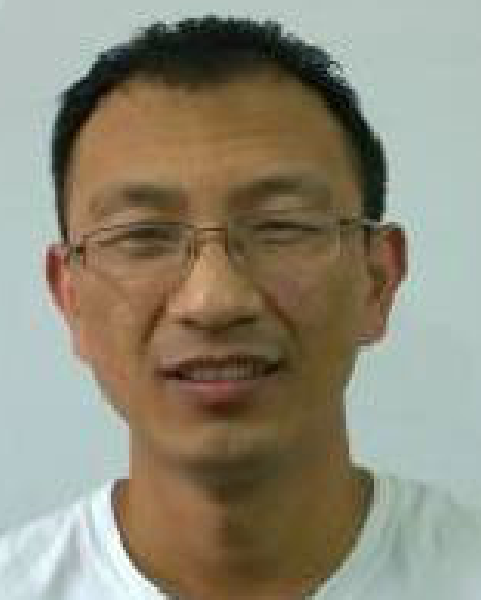}}]{Wangmeng Zuo} (Senior Member, IEEE) received the Ph.D. degree in computer application technology from Harbin Institute of Technology, Harbin, China, in 2007. He is currently a Professor with the Faculty of Computing, Harbin Institute of Technology.
He has published over 200 papers in top tier academic journals and conferences. His current research interests include low level vision, image/video generation, and multimodal understanding. He served as an Associate Editor for IEEE TRANSACTIONS
ON PATTERN ANALYSIS AND MACHINE INTELLIGENCE, IEEE TRANSACTIONS ON IMAGE PROCESSING, and SCIENCE CHINA Information Sciences.
\end{IEEEbiography}

\begin{IEEEbiography}[{\includegraphics[width=1in,height=1.25in,clip,keepaspectratio]{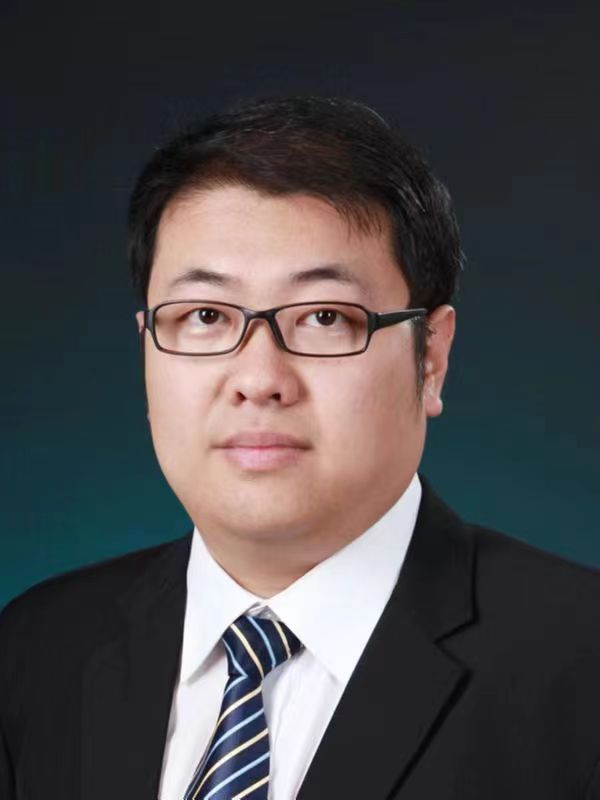}}]{Xiaopeng Fan} (Senior Member, IEEE) received the B.S. and M.S. degrees from the Harbin Institute of Technology (HIT), Harbin, China, in 2001 and 2003, respectively, and the Ph.D. degree from the Hong Kong University of Science and Technology, Hong Kong, in 2009. In 2009, he joined HIT, where he is currently a Professor. From 2003 to 2005, he was with Intel Corporation, China, as a Software Engineer. From 2011 to 2012, he was with Microsoft Research Asia, as a Visiting Researcher. From 2015 to 2016, he was with the Hong Kong University of Science and Technology, as a Research Assistant Professor. He has authored one book and more than 250 articles in refereed journals and conference proceedings. His research interests include video coding and transmission, image processing, and computer vision. He was the Program Chair of PCM2017, Chair of IEEE SGC2015, and Co-Chair of MCSN2015. He was an Associate Editor for IEEE 1857 Standard in 2012. He was the recipient of Outstanding Contributions to the Development of IEEE Standard 1857 by IEEE in 2013.
\end{IEEEbiography}

\begin{IEEEbiography}[{\includegraphics[width=1in,height=1.25in,clip,keepaspectratio]{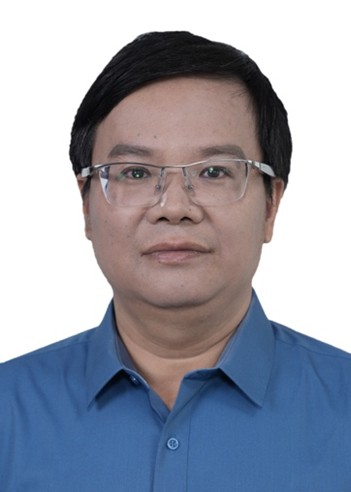}}]{Yonghong Tian} (S’00-M’06-SM’10-F’22) is currently a Boya Distinguished Professor and the Director of Beijing Key Laboratory of Spiking Large Models at the School of Computer Science, and the Vice Chancellor of Shenzhen Graduate School and the Executive Dean of School of AI for Science, Peking University, China. He is also the Deputy Director of Intelligence Supercomputing Division, PengCheng Laboratory, Shenzhen, China. His research interests include neuromorphic computing, distributed machine learning and AI for Science. He is the author or co-author of over 400 technical articles in refereed journals and conferences such as Nature Machine Intelligence, Nature Computational Science, Nature Communications, Science Advances, IEEE TPAMI, IJCV, ICML, NeuIPS, etc. Prof. Tian was a Senior Associate Editor of IEEE TCSVT (2024.1-now), IEEE TMM (2014.8-2018.8), IEEE Multimedia Mag. (2018.1-2022.8). He co-initiated IEEE Int’l Conf. on Multimedia Big Data (BigMM), and served as the TPC Member of more than ten conferences such as CVPR, ICCV, ACM KDD, AAAI, ACM MM and ECCV. He was the recipient of the Chinese National Science Foundation for Distinguished Young Scholars in 2018 and 2024, two National Science and Technology Awards and five ministerial-level awards in China, the 2022 IEEE SA Standards Medallion and SA Emerging Technology Award, and the 2025 IEEE Hans Karlsson Award. He is a Fellow of IEEE, a senior member of CIE and CCF, a member of ACM.

\end{IEEEbiography}

\end{document}